\documentclass[letterpaper]{article}
\usepackage{icaps2026}
\usepackage{times}
\usepackage{helvet}
\usepackage{natbib}
\usepackage{courier}
\usepackage{caption}
\usepackage{graphicx}
\usepackage[hyphens]{url}

\usepackage{bm}
\usepackage{tikz}
\usepackage{xparse}
\usepackage{amsmath}
\usepackage{amsfonts}
\usepackage{enumitem}
\usepackage{multirow}
\usepackage{booktabs}
\usepackage{algorithm}
\usepackage{subcaption}
\usepackage[table]{xcolor}
\usepackage{algpseudocode}
\usetikzlibrary{arrows.meta, positioning, calc, decorations.text}

\DeclareCaptionStyle{ruled}{labelfont=normalfont,labelsep=colon,strut=off}

\graphicspath{{pdfs/}}
\definecolor{Gray}{gray}{0.92}
\definecolor{lightgray}{HTML}{EFEFEF}
\definecolor{nearwhite}{HTML}{FEFEFE}

\newcommand{\rba}{RBA*}
\newcommand{\Node}{\mathcal{P}}
\newcommand{\problem}{\ensuremath{\Delta}\text{-MAPF}}

\DeclareMathOperator*{\argmin}{\arg\!\min}

\DeclareMathOperator*{\maxdist}{d_{\text{max}}}

\makeatletter
\newcommand{\StateWrapped}[1]{\State \parbox[t]{\dimexpr\linewidth-\ALG@thistlm}{\strut #1\strut}}
\makeatother

\title{Risk-Bounded Multi-Agent Visual Navigation via Iterative Risk Allocation}
\author{
    Viraj Parimi\textsuperscript{\rm 1},
    Brian Williams\textsuperscript{\rm 1}
}
\affiliations{
    \textsuperscript{\rm 1}
    Massachusetts Institute of Technology \\
    vparimi@mit.edu, williams@csail.mit.edu
}

\begin{document}

\maketitle

\begin{abstract}
Safe navigation is essential for autonomous systems operating in hazardous environments, especially when multiple agents must coordinate using only high-dimensional visual observations. While recent approaches successfully combine Goal-Conditioned RL (GCRL) for graph construction with Conflict-Based Search (CBS) for planning, they typically rely on deleting edges with high risk before running CBS to enforce safety. This binary strategy is overly conservative, precluding feasible missions that require traversing high-risk regions, even when the aggregate risk is acceptable. To address this, we introduce a framework for Risk-Bounded Multi-Agent Path Finding (\problem{}), where agents share a user-specified global risk budget ($\Delta$). Rather than permanently discarding edges, our framework dynamically distributes per-agent risk budgets ($\delta_i$) during search via an Iterative Risk Allocation (IRA) layer that integrates with a standard CBS planner. We investigate two distribution strategies: a greedy surplus-deficit scheme for rapid feasibility repair, and a market-inspired mechanism that treats risk as a priced resource to guide improved allocation. The market-based mechanism yields a tunable trade-off wherein agents exploit available risk to secure shorter, more efficient paths, but revert to longer, safer detours under tighter budgets. Experiments in complex visual environments show that our dynamic allocation framework achieves higher success rates than baselines and effectively leverages the available safety budget to reduce travel time. Project website: \url{https://rb-visual-mapf-mers.csail.mit.edu}
\end{abstract}

\section{Introduction}

Safe and efficient multi-agent navigation is critical in domains like disaster relief \cite{disaster2}, large-scale inspection \cite{inspection}, and environmental monitoring \cite{env_monitor}, where failures are costly in both resources and potential harm to people or the environment. Existing Multi-Agent Path Finding (MAPF) \cite{mapf_benchmarks} methods range from exhaustive approaches like Conflict-Based Search (CBS) \cite{sharon-cbs}, which provide high-quality solutions but scale poorly, to prioritized schemes like Priority-Based Search (PBS) \cite{ma2018searchingconsistentprioritizationmultiagent}, which trade optimality for scalability. CBS's modular design has enabled many enhancements \cite{mapf4, mapf5, mapf6} extending to lifelong planning \cite{ma2017lifelongmultiagentpathfinding}, information-guided planning \cite{infomapf} and more recently diffusion-guided planning \cite{2024arXiv241003072S, parimi2025diffusionguidedmultiarmmotionplanning}. However, these methods typically assume access to an explicit or implicit graph with valid transitions and known costs, an assumption that breaks down when agents must operate directly from high-dimensional visual observations where the graph structure is unknown.

\begin{figure*}
\centering
\includegraphics[width=\textwidth]{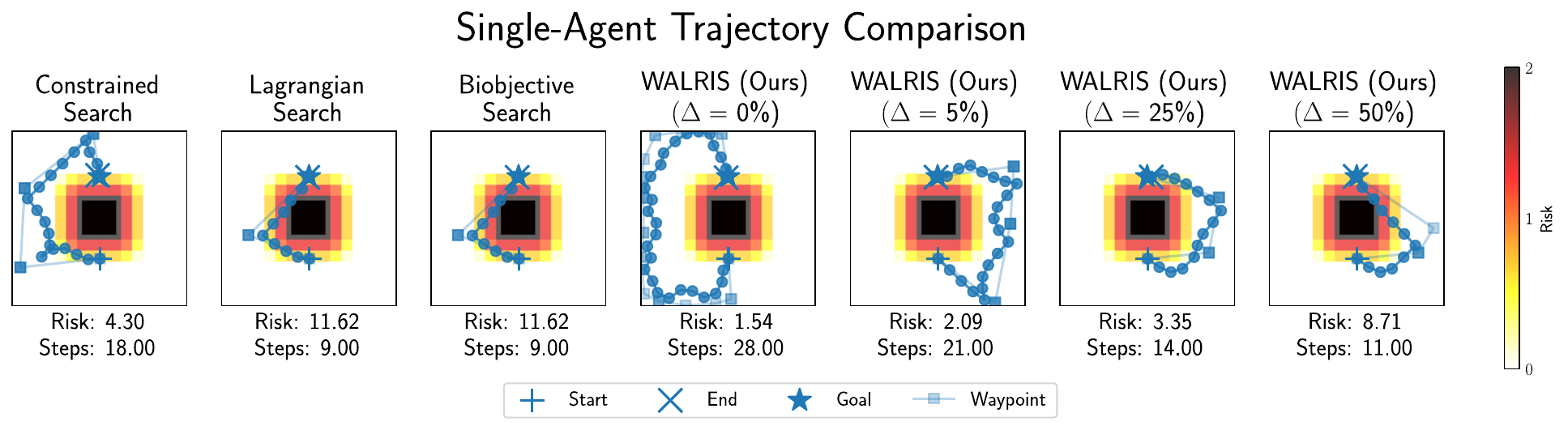}
\caption{Single-agent trajectory comparison on 2D navigation task. Cumulative risk and step count are shown below each plot. Standard baselines are either rigidly conservative (Constrained Search) or overly aggressive (Lagrangian, Biobjective). In contrast, our planner (using the WALRIS strategy) enables a tunable trade-off. As the risk budget $\Delta$ increases, the agent accepts tighter clearances to the hazard to reduce path length, smoothly transitioning from safe detours to efficient direct routes.}
\label{fig:single_pointenv_illustration}
\end{figure*}

\begin{figure*}
    \centering  
    \includegraphics[width=\textwidth]{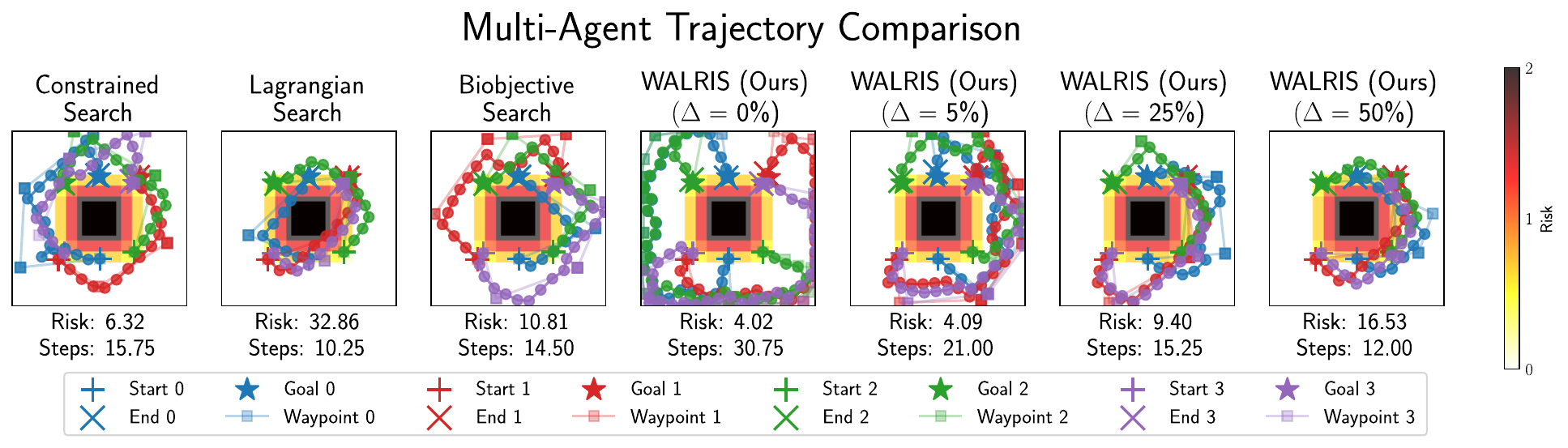}
    \caption{Multi-agent trajectory comparison on the 2D navigation task. Cumulative risk and average step count are annotated below each plot. While baselines are statically locked into specific trade-offs (e.g., Constrained Search is safe but inefficient; Lagrangian is efficient but violates safety), our framework enables a dynamic spectrum of behavior. At strict budgets ($\Delta=0\%$), agents coordinate to take wide, safe detours; as the budget relaxes (e.g., $\Delta=50\%$), they collectively ``spend'' the risk resource to cut through the center, significantly reducing travel time.}
    \label{fig:multi_pointenv_illustration}
\end{figure*}

Goal-conditioned reinforcement learning (GCRL) \cite{mirowski2017learning, pmlr-v37-schaul15,pong2020temporaldifferencemodelsmodelfree} excels at learning a single navigation policy directly from complex observations across many goals, making it well suited for visually rich and unstructured environments. However, GCRL alone often struggles on long-horizon tasks \cite{levy2019hierarchical,nachum2018dataefficient}, especially when balancing risk against efficiency. While Constrained RL \cite{altman2021constrained}, Control Barrier Functions (CBFs) and Safe MARL \cite{cbf_mapf,cbf_safe_rl_1,cbf_safe_rl_2} offer robust tools for enforcing safety, they typically operate reactively by shaping low-level actions to satisfy hard state constraints. In addition, recent hybrid approaches \cite{eysenbach2019search, feng2025safemultiagentnavigationguided} integrate GCRL with graph-based planning. These methods build an intermediate waypoint graph from a replay buffer, learn distance and risk critics, prune edges deemed unsafe, and then apply CBS to coordinate multiple agents. This approach yields safer waypoint-based plans that respect the learned safety critics. However, such static edge pruning is fundamentally binary, as edges exceeding a local threshold are permanently discarded. This rigidity proves to be overly conservative in missions where goals require entering hazardous regions, or when accepting a small, controlled amount of risk could dramatically reduce travel time or even make an otherwise infeasible mission possible (Figs.~\ref{fig:single_pointenv_illustration}--\ref{fig:single_habitatenv_illustration}). Figures~\ref{fig:single_pointenv_illustration}--\ref{fig:multi_pointenv_illustration} show representative 2D examples (single- and multi-agent), while Figure~\ref{fig:single_habitatenv_illustration} shows the analogous effect in a visually complex environment. This is a challenge that demands \textit{budgeted risk acceptance} rather than strict avoidance. 

Parallel to these developments, risk allocation in optimal control methods like Iterative Risk Allocation (IRA) \cite{ono2008iterative} and its extensions, such as MIRA \cite{ono2012robust, ono2010market}, treat a global chance constraint as a shared resource. By redistributing risk from ``easy'' constraints to ``hard'' ones, they mitigate the conservatism of uniform risk allocations. However, these methods rely on convex trajectory optimization in continuous-state spaces and do not address the discrete, combinatorial structure of multi-agent conflict resolution, nor do they handle planning over unstructured and learned graphs.

To address these limitations, we introduce the \emph{Risk-Bounded Multi-Agent Path Finding} (\problem{}) framework, where all agents share a user-specified global risk budget $\Delta$. Instead of statically pruning unsafe edges, we augment the standard CBS constraint tree with an IRA layer that maintains per-agent budgets $\boldsymbol{\delta}$ and redistributes them upon node infeasibility. Within this framework, we investigate two complementary risk distribution strategies grounded in economic principles. The first, \emph{EQUIRIS}, is a greedy surplus-deficit scheme that shifts risk in an equity-like fashion from agents with slack to those in need, serving as a fast feasibility repair heuristic. The second, \emph{WALRIS}, is a Walrasian tatonnement-inspired mechanism that treats risk as a priced resource and lets agents independently trade off path length against risk at a shared price signal. By varying the global budget $\Delta$, WALRIS yields a smooth, user-controllable trade-off. When the global budget is generous, agents exploit higher local risk allowances to find shorter paths, while tighter budgets induce longer but safer detours aligning behavior with user-defined safety preferences.

In summary, our contributions are threefold. First, we formalize the \problem{} problem on learned waypoint graphs subject to a global risk bound $\Delta$. Second, we augment the CBS constraint tree with a discrete Iterative Risk Allocation (IRA) layer that dynamically redistributes per-agent budgets via two complementary strategies, EQUIRIS or WALRIS. Finally, we demonstrate superior safety-efficiency trade-offs over baselines in both 2D and complex visual environments, as well as a ROS2/Gazebo integration controlling multiple Crazyflie drones in simulation and hardware.

The remainder of the paper is organized as follows. Section~\ref{sec:prelims} establishes the background on GCRL, learned waypoint graphs, and the global risk formulation. Section~\ref{sec:approach} details the \problem{} framework, introducing our two redistribution strategies, EQUIRIS and WALRIS. Section~\ref{sec:experiments} presents experimental results evaluating adaptability, success rates, and scalability across 2D and complex visual environments. Finally, we conclude in Section~\ref{sec:conclusion}.

\section{Preliminaries}
\label{sec:prelims}

We consider multi-agent navigation in visually rich environments, where each agent must reach a goal while ensuring that the \emph{total accumulated risk} across all agents does not exceed a global risk bound $\Delta$.

\subsection{Goal-Conditioned Reinforcement Learning}

In GCRL, an agent interacts with an environment modeled as a Markov Decision Process (MDP) $(\mathcal{S}, A, P, R, \gamma)$ augmented with a goal $s_g \in \mathcal{S}$. Here \(\mathcal{S}\) denotes the high-dimensional state space, \(A\) is the action space, \(P(s_{t+1} \mid s_t, a_t)\) is the transition dynamics, \(R(s, a, s_g)\) is a goal-dependent reward providing feedback on progress towards $s_g$, and \(\gamma \in [0, 1)\) is the discount factor. The agent learns a policy \(\mu_{\theta}(a \mid s, s_g)\) that maximizes the expected cumulative reward conditioned on both the current state \(s\) and goal \(s_g\). While GCRL is effective for learning short-horizon skills, often leveraging reward shaping \cite{chiang2019learning} or demonstrations \cite{lynch2019learning,nair2018overcoming}, we use it here to abstract low-level visual control into a navigable graph via learned distance and risk critics.

\subsection{Learned Distance and Risk Graph}

Following \citeauthor{feng2025safemultiagentnavigationguided}, we employ a dual critic architecture training two value functions, $\mathcal{Q}^d_{\theta}(s, a, s_g)$ and $\mathcal{Q}^c_{\theta}(s, a, s_g)$, to estimate the distance and risk between state-goal pairs. For a pair of states $(s_i, s_j)$ and action $a_{ij} = \mu_{\theta}(s_i, s_j)$ derived from the policy, we define the metrics 
\begin{align*}
    d_\mu(s_i, s_j) &\gets \mathcal{Q}^d_{\theta}(s_i, a_{ij}, s_j), \\
    c_\mu(s_i, s_j) &\gets \mathcal{Q}^c_{\theta}(s_i, a_{ij}, s_j).
\end{align*}
where $d_{\mu}$ approximates shortest-path distance and $c_{\mu}$ estimates the cumulative risk of traversing from $s_i$ to $s_j$. From the agent’s replay buffer, we sample a set of states $\mathcal{B}$ and build a directed graph $\mathcal{G} = (\mathcal{V}, \mathcal{E}, \mathcal{W}_d, \mathcal{W}_c)$ with
\begin{align*}
    \mathcal{V} &= \mathcal{B}, \\
    \mathcal{E} &= \{ e_{s_i \rightarrow s_j} \mid s_i, s_j \in \mathcal{B} \}, \\
    \mathcal{W}_d(e_{s_i \rightarrow s_j}) &=
        \begin{cases}
            d_\mu(s_i, s_j), & \text{if } d_\mu(s_i, s_j) < \maxdist, \\
            \infty, & \text{otherwise},
        \end{cases} \\
    \mathcal{W}_c(e_{s_i \rightarrow s_j}) &= c_\mu(s_i, s_j).
\end{align*}

Unlike \citeauthor{feng2025safemultiagentnavigationguided}, we do not prune edges based on predicted risk. All edges with finite distances are retained, including those that may pass through hazardous regions. The risk critic $\mathcal{Q}^c_{\theta}$ instead provides additional risk information to the planner, which is used to reason about trade-offs under the global risk bound $\Delta$. 

\subsection{Global Risk-Bounded MAPF}

Let $\mathcal{A} = \{a_1, \dots, a_N\}$ be a set of $N$ agents, each with a start-goal pair $(s_i, g_i) \in \mathcal{V} \times \mathcal{V}$. Our objective is to find a joint plan $\Pi = \{\pi_1, \dots, \pi_N\}$ of collision-free paths, where each $\pi_i$ connects $s_i$ to $g_i$, such that the total accumulated risk does not exceed a user-specified global budget $\Delta$. We define the length of a single path $\pi_i$ as $\ell(\pi_i) = \sum_{e \in \pi_i} \mathcal{W}_d(e)$, where $\mathcal{W}_d(e)$ is the learned distance cost associated with edge $e$. We then define the risk of a single path $\pi_i$ as $\rho(\pi_i) = \sum_{e \in \pi_i} \mathcal{W}_c(e)$, where $\mathcal{W}_c(e)$ is the learned risk associated with edge $e$. The \problem{} problem seeks to minimize the sum of path lengths, $\mathcal{J}(\Pi) = \sum_{i=1}^N \ell(\pi_i)$ subject to the global risk constraint $\sum_{i=1}^N \rho(\pi_i) \;\leq\; \Delta$.

This formulation treats risk as a \textit{shared resource}, enabling the system to handle heterogeneous environments where tasks naturally vary in difficulty. For instance, in a search-and-rescue scenario, one agent may need to enter a hazardous zone while others remain in safe areas. Enforcing uniform per-agent risk limits would likely render such missions infeasible. Instead, by pooling the budget, our approach allows some agents to draw a larger share of risk when necessary for mission success. Furthermore, $\Delta$ serves as a single control knob that can adjust the degree of the imbalance. If a resulting plan is deemed too aggressive for a single agent, the user can tighten $\Delta$, shrinking the available resource pool and forcing the planner to redistribute risk, yielding more conservative and balanced solutions.

\section{Approach}
\label{sec:approach}

We address the \problem{} problem by augmenting a standard CBS planner with an \emph{Iterative Risk Allocation} (IRA) layer. While standard CBS focuses solely on resolving spatio-temporal conflicts, our planner simultaneously manages the distribution of the global risk budget $\Delta$. To do so, we explicitly maintain a vector of local risk budgets $\boldsymbol{\delta} = [\delta_1, \dots, \delta_N]$ within the high-level search nodes allowing the planner to dynamically redistribute risk among agents when local constraints become too tight. Inspired by iterative risk allocation methods \cite{ono2008iterative, ono2012robust}, we adapt this concept here as a discrete, reallocation step. Whenever a node becomes infeasible under its current budgets, we adjust $\boldsymbol{\delta}$ and continue the search in that branch. Upon finding a valid joint plan, agents execute the generated waypoints using the underlying goal-conditioned policy.

\subsection{High-Level Search}

The high-level search explores a Constraint Tree (CT), as in standard CBS, but augments each node with a risk-allocation state. A node is defined as a tuple $\Node = (\mathcal{C}, \Pi, \boldsymbol{\delta}, \boldsymbol{\phi}, \mathcal{J})$ containing the spatio-temporal constraints $\mathcal{C}$, the current set of single-agent paths $\Pi = \{\pi_1, \ldots, \pi_N\}$, and the sum-of-costs objective $\mathcal{J} = \mathcal{J}(\Pi) = \sum_i \ell(\pi_i)$. Additionally, we store the local risk budget vector, $\bm{\delta} = [\delta_1, \dots, \delta_N]$, and a boolean validity vector $\bm{\phi} \in \{0, 1\}^N$. Here, $\phi_i = 1$ indicates that a valid path for agent $a_i$ is stored in $\Pi$ satisfying both $\mathcal{C}$ and $\delta_i$ has been found, while $\phi_i = 0$ implies otherwise. 

\begin{algorithm}[!ht]
\caption{\problem{}: High-Level Search}
\label{alg:rbmapf}
\begin{algorithmic}[1]
\State Create root node $\Node_0$ with initial (unconstrained) paths $\Pi^0$, and risk allocations $\boldsymbol{\delta}^0$.
\State $\mathcal{F}.\texttt{Insert}(\Node_0)$
\While{$\mathcal{F}$ not empty and time not exceeded}
    \StateWrapped{$\Node \gets \mathcal{F}.\texttt{ExtractMin()}$ \Comment{\textit{Pop node from the frontier set with lowest priority key}}}
    \State $\Pi \gets \Node.\Pi, \; \boldsymbol{\delta} \gets \Node.\boldsymbol{\delta}, \; \boldsymbol{\phi} \gets \Node.\boldsymbol{\phi}$
    \If{$\exists i \text{ s.t. } \phi_i = 0$} \label{alg:phase-1-start}
        \State $\mathcal{A}_{\text{fail}} \gets \emptyset$
        \For{each $a_i$ with $\phi_i = 0$}
            \State $(\pi_i, \ell_i, \rho_i) \gets \textsc{\rba{}}(a_i, \Node, \delta_i)$
            \If{\texttt{FAIL}} 
                \State $\mathcal{A}_{\text{fail}} \gets \mathcal{A}_{\text{fail}} \cup \{a_i\} $
            \Else
                \State Update $\Pi$ with $\pi_i$ and set $\phi_i \gets 1$
            \EndIf
        \EndFor
        \If{$\mathcal{A}_{\text{fail}} \neq \emptyset$}
            \State $\boldsymbol{\delta}' \gets \textsc{ReallocateRisk}(\Node, \mathcal{A}_{\text{fail}})$
            \If{$\boldsymbol{\delta}' \neq \texttt{FAIL}$}
                \State $\Node' \gets$ copy of $\Node, \; \Node'.\boldsymbol{\delta} \gets \boldsymbol{\delta}'$
                \State $\text{Update } \Node'.\phi_j \gets 0 \; \forall a_j \in \mathcal{A}_{\text{fail}}$
                \State $\mathcal{F}.\texttt{Insert}(\Node')$
            \EndIf
            \State \textbf{continue}
        \EndIf
    \EndIf \label{alg:phase-1-end}
    \State $\mathcal{K} \gets \textsc{DetectCollisions}(\Pi)$ \label{alg:phase-2-start}
    \If{$\mathcal{K} = \emptyset$ and $\sum_{i=1}^N \rho(\pi_i) \le \Delta$}
        \State \Return $\Pi$ \Comment{\textit{Solution found}}
    \EndIf
    \State $c \gets \textsc{SelectCollision}(\mathcal{K})$
    \StateWrapped{Generate disjoint split constraints for collision $c$}
    \For{each new constraint on agent $a_k$}
        \StateWrapped{$\Node' \gets$ copy of $\Node$ with added constraint}
        \State $(\pi_k', \ell_k', \rho_k') \gets \textsc{\rba{}}(a_k, \Node', \delta_k)$
        \If{\texttt{FAIL}}
            \State $\boldsymbol{\delta}' \gets \textsc{ReallocateRisk}(\Node', \{a_k\})$
            \If{$\boldsymbol{\delta}' \neq \texttt{FAIL}$}
                \State $\Node'.\boldsymbol{\delta} \gets \boldsymbol{\delta}', \; \Node'.\phi_k \gets 0$
                \State $\mathcal{F}.\texttt{Insert}(\Node')$
            \EndIf
        \Else
            \State Update $\Node'.\Pi$ with $\pi_k'$
            \StateWrapped{$\Node'.\mathcal{J} \gets \textsc{SumOfCosts}(\Node'.\Pi), \Node'.\phi_k \gets 1$}
            \State $\mathcal{F}.\texttt{Insert}(\Node')$
        \EndIf
    \EndFor \label{alg:phase-2-end}
\EndWhile
\State \Return No solution found.
\end{algorithmic}
\end{algorithm}

The high-level search (Algorithm~\ref{alg:rbmapf}) maintains a priority queue $\mathcal{F}$ of CT nodes. Nodes are prioritized primarily by the sum-of-costs $\mathcal{J}(\Pi)$, breaking ties with the number of unresolved collisions and, finally, the number of recent risk reallocations. This favors nodes that are both low-cost and stable in their risk distributions. The main loop begins by computing initial single-agent paths on the learned waypoint graph (ignoring risk) and an initial risk allocation $\boldsymbol{\delta}^0$ (Section~\ref{subsec:init_alloc}). 

Each iteration proceeds in two phases. In \emph{Phase~1} (lines~\ref{alg:phase-1-start}-\ref{alg:phase-1-end}), we attempt to compute valid paths for any agents marked as invalid $(\phi_i = 0)$ using the risk-constrained \rba{} planner (Section~\ref{sec:rba}). If any agent fails to find a path within its budget $\delta_i$, the set of failing agents is collected, and a risk allocation step is attempted. If successful, a new child node with the updated budgets is added to $\mathcal{F}$ while an unsuccessful reallocation causes the current node to be pruned. 

In \emph{Phase~2} (lines~\ref{alg:phase-2-start}-\ref{alg:phase-2-end}), once all agents have valid paths, the search reduces to standard CBS. We detect collisions in $\Pi$, return a solution if none remain and $\sum_i \rho(\pi_i) \le \Delta$ is satisfied. Otherwise, we branch on a selected collision using disjoint split \cite{LiHS0K19}. For each child, we attempt to replan the newly constrained agent with its current budget. Crucially, if this replanning fails due to the new constraint, we immediately trigger the risk allocation layer to determine whether budget adjustment can restore feasibility. In this way, the CT search and the risk-allocation layer interact tightly where CBS handles discrete conflicts, while the allocation layer reshapes $\boldsymbol{\delta}$ to keep promising branches feasible under the global risk bound.

\subsection{Low-Level Search}
\label{sec:rba}

The low-level single-agent planner is a risk-bounded variant of A*, denoted \rba{}. Given an agent $a_i$, a set of constraints from node $\Node$, and a budget $\delta_i$, it searches the learned waypoint graph for a path $\pi_i$ that minimizes length $\ell(\pi_i)$ while satisfying $\rho(\pi_i) \leq \delta_i$. To ensure soundness, \rba{} operates on an augmented state space $(u, r)$, where $u \in \mathcal{V}$ is the current node and $r$ is the accumulated risk. During search, we prune any state where $r > \delta_i$. Crucially, we employ dominance pruning \cite{moa} where a path to node $u$ with length $\ell$ and risk $\rho$ is discarded if and only if there exists a previously discovered path to $u$ with length $\ell' \leq \ell$ and risk $\rho' \leq \rho$. Formally, the planner returns
\[
    \pi_i^*(\delta_i) \in \argmin_{\pi_i \in \Pi_{\Node} : \rho(\pi_i) \leq \delta_i} \ell(\pi_i).
\]
where $\Pi_{\Node}$ is the set of paths satisfying the spatio-temporal constraints in $\Node$.

Besides \textsc{\rba{}}, the risk allocation layer uses two simpler unconstrained A*-based queries on the same learned waypoint graph at a CT node $\Node$.

\textbf{Minimum feasible risk} \(\textsc{MinFeasibleRisk}(a_i, \Node)\).  
This query finds the safest path regardless of length. We run A* using learned risk weights $\mathcal{W}_c$ and a risk-based heuristic to find $\pi_i^{\text{risk}} \in \argmin_{\pi_i \in \Pi_{\Node}} \rho(\pi_i)$, yielding the lower bound $\delta_i^{\min} \gets \rho(\pi_i^{\text{risk}})$

\textbf{Risk of shortest path} \(\textsc{LenMinRisk}(a_i, \Node)\).  
This query finds the shortest path regardless of risk. We run A* using the learned distance weights $\mathcal{W}_d$ to find, $\pi_i^{\text{len}} \in \argmin_{\pi_i \in  \Pi_{\Node}} \ell(\pi_i)$, yielding the upper bound $\delta_i^{\max} \gets \rho(\pi_i^{\text{len}})$.

\begin{figure*}
    \centering
    \includegraphics[width=\textwidth]{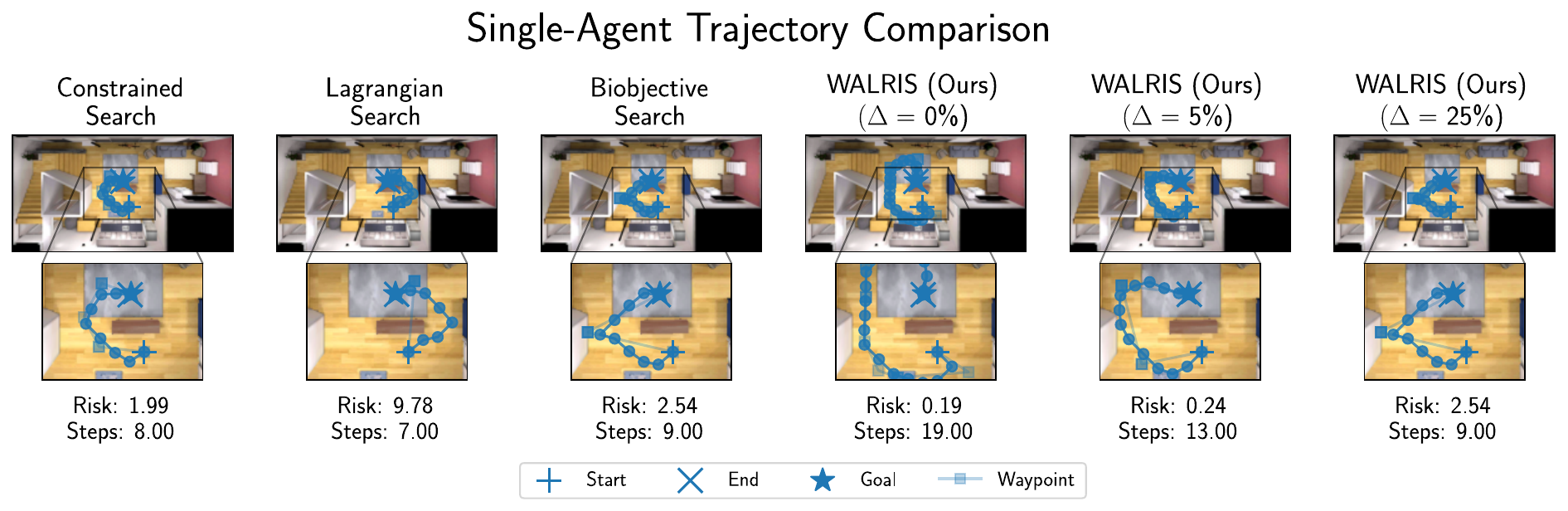}
    \caption{Single-agent trajectory comparison on the visual navigation task. Cumulative risk and step count are shown below each plot. While baselines produce static solutions that can be excessively risky (Lagrangian) or rigid, our planner enables a dynamic trade-off controlled by $\Delta$. At strict budgets ($\Delta=0\%$), the agent takes a significant detour to ensure maximal safety, smoothly transitioning to direct, efficient routes as the allowed risk budget increases.}
    \label{fig:single_habitatenv_illustration}
\end{figure*}

\subsection{Risk Budget Management}
\label{subsec:risk_realloc}

The core innovation of our approach is its explicit management of per-agent risk budgets under the global constraint $\Delta$. This has two components: (i) determining an initial allocation at the root, and (ii) iterative reallocation strategy when agents fail to find feasible paths.

\subsubsection{Initial Budget Allocation}
\label{subsec:init_alloc}

At the root of the CT, we compute an initial allocation $\boldsymbol{\delta}^0$ based on metrics derived from the agents’ unconstrained paths on the learned waypoint graph. We use three schemes defined by a utility term $w_i$:
\begin{itemize}
    \item \textbf{Uniform}: Divide the budget equally, $\delta_i = \Delta / N$.
    \item \textbf{Utility-based}: $\delta_i = \Delta \cdot \frac{w_i}{\sum_{j} w_j}$, e.g., $w_i = \rho_i$ to give more budget to agents with riskier unconstrained paths.
    \item \textbf{Inverse utility-based}: $\delta_i = \Delta \cdot \frac{1/w_i}{\sum_{j} (1/w_j)}$, e.g., $w_i = \ell_i$ to favor shorter, potentially riskier paths.
\end{itemize}
Following initialization, the search invokes the IRA layer to adjust $\boldsymbol{\delta}$ whenever a node becomes infeasible. To do so, we introduce two economically inspired strategies: \emph{EQUIRIS}, a surplus-deficit scheme that transfers risk from agents with slack to those in need of risk in an equity-like fashion, and \emph{WALRIS}, a Walrasian tatonnement-inspired \cite{tuinstra2012price} mechanism that treats risk as a traded resource balancing global demand against the budget $\Delta$.

\begin{algorithm}[!ht]
\caption{EQUIRIS Strategy}
\label{alg:reallocate_risk_sd}
\begin{algorithmic}[1]
\Function{ReallocateRisk}{$\Node$, $\mathcal{A}_{\text{fail}}$}
    \State $\delta_{\text{req}} \gets 0$
    \For{each failing agent $a_i \in \mathcal{A}_{\text{fail}}$} \label{alg:required-start}
        \StateWrapped{$\delta^{\min}_i \gets \textsc{MinFeasibleRisk}(a_i, \Node)$}
        \State $\delta_{\text{req}} \gets \delta_{\text{req}} + (\delta^{\min}_i - \Node.\delta_i)$
    \EndFor \label{alg:required-end}
    
    \State $\delta_{\text{avail}} \gets 0$ 
    \For{each passing agent $a_j \notin \mathcal{A}_{\text{fail}}$} \label{alg:surplus-start}
        \StateWrapped{$\delta^{\min}_j \gets  \textsc{MinFeasibleRisk}(a_j, \Node)$}
        \State $\delta_{\text{avail}} \gets \delta_{\text{avail}} + (\Node.\delta_j - \delta^{\min}_j)$
    \EndFor \label{alg:surplus-end}
    
    \If{$\delta_{\text{req}} > \delta_{\text{avail}}$}
        \StateWrapped{\Return \texttt{FAIL} \Comment{\textit{Not enough surplus budget}}}
    \EndIf
    
    \State $\boldsymbol{\delta}' \gets \Node.\boldsymbol{\delta}$ \label{alg:reallocate-start}
    \For{each failing agent $a_i \in \mathcal{A}_{\text{fail}}$}
        \StateWrapped{$\delta'_i \gets \delta^{\min}_i$ \Comment{\textit{Assign minimum required}}}
    \EndFor
    
    \StateWrapped{$\delta_{\text{rem}} \gets \delta_{\text{req}}$ \Comment{\textit{Deduct from surplus agents}}}
    \For{each passing agent $a_j \notin \mathcal{A}_{\text{fail}}$} 
        \State $\Delta \delta_j \gets \Node.\delta_j - \delta^{\min}_j$
        \State $\epsilon_j \gets \min(\Delta \delta_j, \delta_{\text{rem}})$
        \State $\delta'_j \gets \delta'_j - \epsilon_j$
        \State $\delta_{\text{rem}} \gets \delta_{\text{rem}} - \epsilon_j$
        \If{$\delta_{\text{rem}} \le 0$} \textbf{break} \EndIf
    \EndFor \label{alg:reallocate-end}
    
    \State \Return $\boldsymbol{\delta}'$
\EndFunction
\end{algorithmic}
\end{algorithm}

\subsubsection{EQUItable RIsk-bounded Search (EQUIRIS)}

EQUIRIS is a greedy surplus-deficit scheme designed to ``rescue'' infeasible nodes by transferring risk budget from agents with slack to those in deficit. As outlined in Algorithm~\ref{alg:reallocate_risk_sd}, EQUIRIS proceeds in three steps. First (lines~\ref{alg:required-start}-\ref{alg:required-end}), it quantifies the total deficit $\delta_{\text{req}}$ by summing the difference between the minimum required risk $\delta^{\min}_i$ (computed via \textsc{MinFeasibleRisk}) and the current budget $\delta_i$ for all failing agents. Second (lines~\ref{alg:surplus-start}-\ref{alg:surplus-end}), it calculates the total available surplus $\delta_{\text{avail}}$ from the passing agents. If $\delta_{\text{req}} > \delta_{\text{avail}}$, even the most generous redistribution cannot satisfy all failing agents, so the node is declared infeasible and pruned. Finally (lines~\ref{alg:reallocate-start}-\ref{alg:reallocate-end}), we construct the new allocation $\boldsymbol{\delta}'$ by assigning each failing agent its minimum requirement $\delta_i^{\min}$ and greedily deducting the balance from the passing agents’ surpluses until the deficit is covered. 

EQUIRIS is computationally efficient, relying solely on minimum-risk queries to rebalance risk distribution. However, it does not explicitly reason about the length-risk trade-off. By focusing only on feasibility repair, it may miss reallocations that could yield superior sum-of-costs objectives.

\begin{algorithm}[!ht]
\caption{WALRIS Strategy}
\label{alg:reallocate_risk_market}
\begin{algorithmic}[1]
\Function{ReallocateRisk}{$\Node$, $\mathcal{A}_{\text{fail}}$}
    \StateWrapped{$\boldsymbol{\delta} \gets \Node.\boldsymbol{\delta}$ \Comment{\textit{current allocation is starting point}}}
    \For{each agent $a_i \in \mathcal{A}$}
        \StateWrapped{$\delta^{\min}_i \gets \textsc{MinFeasibleRisk}(a_i, \Node)$}
        \StateWrapped{$\delta^{\max}_i \gets \textsc{LenMinRisk}(a_i, \Node)$}
    \EndFor
    \If{$\sum_i \delta^{\min}_i > \Delta$} \Return \texttt{FAIL} \EndIf
    \If{$\sum_i \delta^{\max}_i \leq \Delta$} \Return $\{\delta^{\max}_i\}_{i=1}^N$ \EndIf
    \StateWrapped{$(p_{\min}, p_{\max}) \gets \textsc{InitPriceBounds}(\{\delta^{\min}_i\}, \{\delta^{\max}_i\})$}
    \State $\mathcal{J}^* \gets \infty, \; \boldsymbol{\delta}^{\text{best}} \gets \texttt{None}, \; k \gets 0$
    \While{$p_{\max} - p_{\min} \geq \epsilon$ \textbf{and} $k < K_{\max}$} 
        \State $p \gets (p_{\min} + p_{\max}) / 2$ \label{alg:price-start}
        \For{each agent $a_i \in \mathcal{A}$}
            \StateWrapped{Define a discrete neighborhood set $\mathcal{N}_i$ around $\delta_i$ clipped to $[\delta^{\min}_i, \delta^{\max}_i]$}
            \State $s^*_i \gets \infty$
            \For{each $\hat{\delta} \in \mathcal{N}_i$}
                \StateWrapped{$(\pi_i, \ell_i, \rho_i) \gets \textsc{\rba{}}(a_i, \Node, \hat{\delta})$}
                \If{\texttt{FAIL}} \textbf{continue} \EndIf
                \State $s_i \gets \ell_i + p \cdot \rho_i$
                \If{$s_i < s^*_i$}
                    \StateWrapped{$s^*_i \gets s_i; \delta_i^{\text{new}} \gets \hat{\delta}; \pi_i^{\text{new}} \gets \pi_i$}
                \EndIf
            \EndFor
            \If{$s^*_i = \infty$} \Return \texttt{FAIL} \EndIf
            \State $\delta_i \gets \delta_i^{\text{new}}, \; \text{Update } \Pi \text{ with } \pi_i^{\text{new}}$
        \EndFor \label{alg:price-end}
        \State $\mathcal{J}(\Pi) \gets \textsc{SumOfCosts}(\Pi)$
        \If{$\sum_i \rho(\pi_i) \le \Delta$}
            \If{$\mathcal{J}(\Pi) < \mathcal{J}^*$}
                \State $\mathcal{J}^* \gets \mathcal{J}(\Pi)$; $\boldsymbol{\delta}^{\text{best}} \gets \boldsymbol{\delta}$
            \EndIf
            \StateWrapped{$p_{\max} \gets p$ \Comment{\textit{risk underused; decrease upper price bound}}}
        \Else
            \StateWrapped{$p_{\min} \gets p$ \Comment{\textit{risk overused; increase lower price bound}}}
        \EndIf
        \State $k \gets k + 1$
    \EndWhile
    \If{$\boldsymbol{\delta}^{\text{best}} = \texttt{None}$} \Return \texttt{FAIL} \EndIf
    \State \Return $\boldsymbol{\delta}^{\text{best}}$
\EndFunction
\end{algorithmic}
\end{algorithm}

\subsubsection{WALrasian RIsk-bounded Search (WALRIS)}

While EQUIRIS efficiently repairs feasibility, its fixed ordering of surplus donors limits its ability to improve the global objective. It may overlook allocations where drawing surplus from a different subset of agents would yield a lower total path length. WALRIS addresses this by introducing a market-based allocator inspired by Walrasian tatonnement \cite{tuinstra2012price, ono2010market}. This method treats risk as a scarce, priced resource. Instead of prescribing rigid transfers, WALRIS broadcasts a scalar \emph{price of risk} $p \geq 0$, allowing each agent to independently improve its local trade-off between path length and risk. 

As detailed in Algorithm~\ref{alg:reallocate_risk_market}, WALRIS first computes the feasible risk range $[\delta_i^{\min}, \delta_i^{\max}]$ for each agent. If $\sum_i \delta_i^{\min} > \Delta$, no feasible allocation exists and the node is pruned. Conversely, if $\sum_i \delta_i^{\max} \leq \Delta$, the budget is sufficient for all agents to take their length-optimal paths. In the non-trivial case where budget enforces a trade-off, we initialize a price interval $[p_{\min}, p_{\max}]$ and search for a ``clearing'' price. We defer the initialization of this interval to the Supplement Sec. B.2. Inside the optimization loop (lines~\ref{alg:price-start}-\ref{alg:price-end}), given a candidate price $p$, each agent $a_i$ explores a discrete neighborhood $\mathcal{N}_i$ around its current budget. For each candidate $\hat{\delta} \in \mathcal{N}_i$, the agent computes the path using \rba{} and selects the path that minimizes the price augmented objective 
\[
    s_i(\hat{\delta}, p) = \ell(\pi^*_i(\hat{\delta})) + p\,\cdot\rho(\pi_i^*(\hat{\delta})).
\]

After collecting responses, WALRIS aggregates the total risk $\sum_i \rho(\pi_i)$ and the sum-of-costs $\mathcal{J}(\Pi)$. If $\sum_i \rho(\pi_i) \le \Delta$, the allocation is feasible. We record it if it improves the best known $\mathcal{J}(\Pi)$ and then \emph{decrease} $p_{\max}$ to lower the cost of risk, encouraging agents to find shorter, riskier paths. Conversely, if $\sum_i \rho(\pi_i) > \Delta$, we \emph{increase} $p_{\min}$ to make risk more expensive, forcing agents towards safer, longer paths. This bisection process continues until convergence or a budget of iterations is exhausted. By enabling agents to ``buy'' risk based on their marginal utility, WALRIS achieves a more globally coordinated and efficient distribution of the safety budget than greedy methods. Since the allocation layers presented are heuristic in nature, we include a discussion of the resulting optimality and completeness properties in Supplement Sec. B.1.

\begin{figure*}
    \centering
    \includegraphics[width=\textwidth]{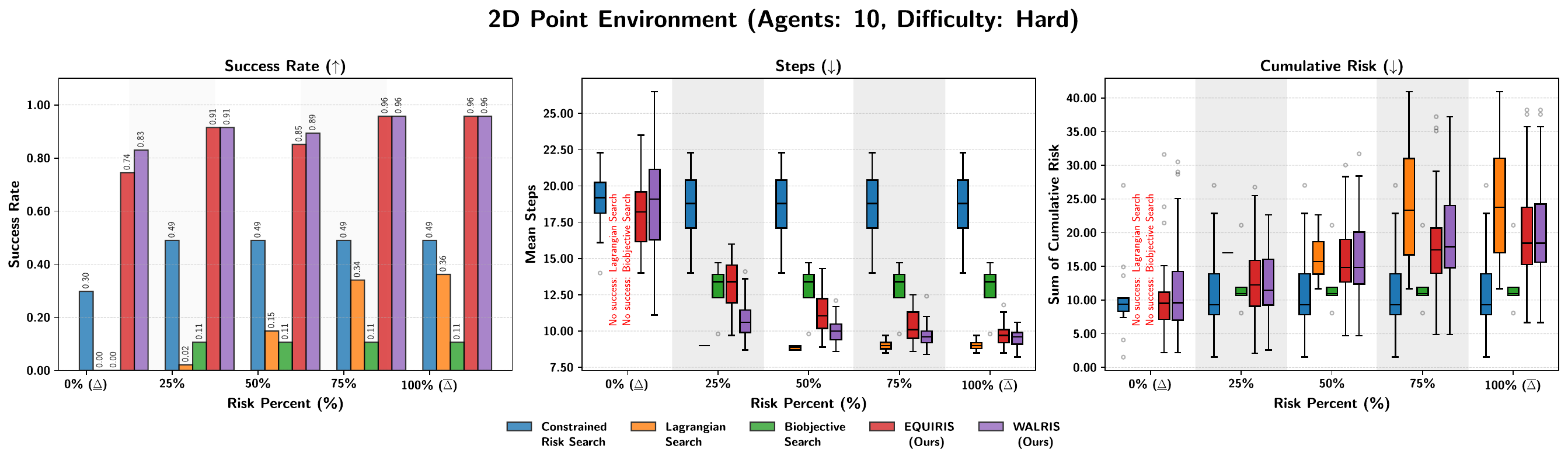}
    \caption{Quantitative performance on the 2D point environment (10 agents, Hard difficulty) as a function of the risk budget $\Delta$. \textbf{Left:} Success rate. \textbf{Center:} Average steps (conditioned on success). \textbf{Right:} Total cumulative risk. \textbf{EQUIRIS} and \textbf{WALRIS} (Ours) maintain high success rates at strict budgets ($\Delta=0\%$) where Lagrangian and Biobjective baselines fail. As $\Delta$ increases (moving right), our planner reduces travel time (Center), smoothly transitioning from safe detours to efficient trajectories.}
    \label{fig:pointenv_plot}
\end{figure*}

\begin{figure*}[t]
    \centering
    \includegraphics[width=\textwidth]{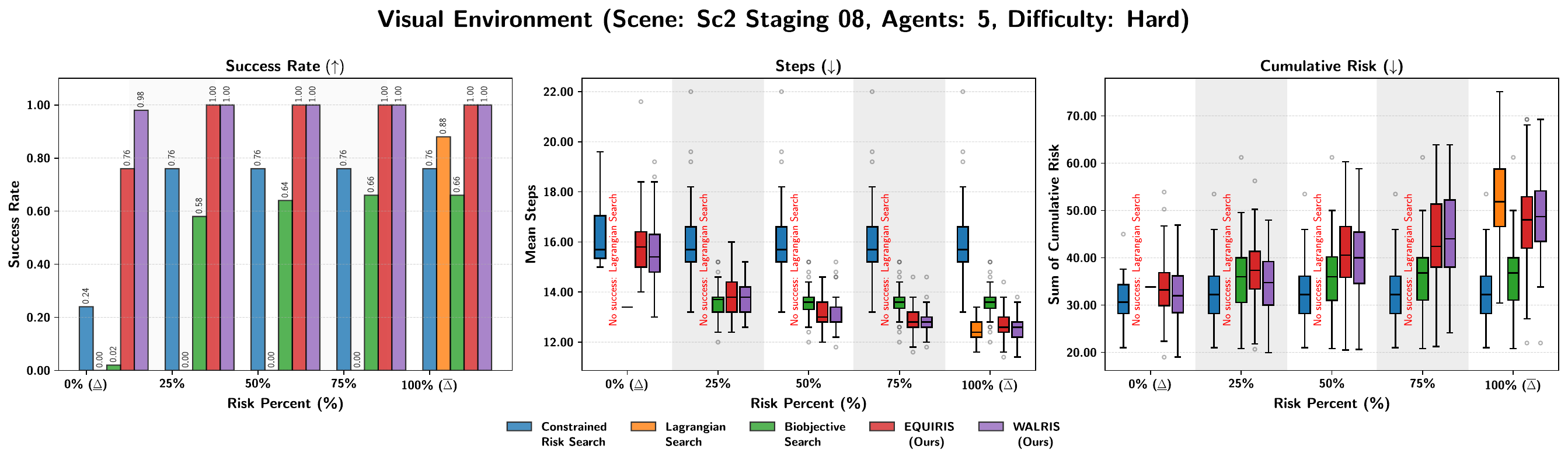}
    \includegraphics[width=\textwidth]{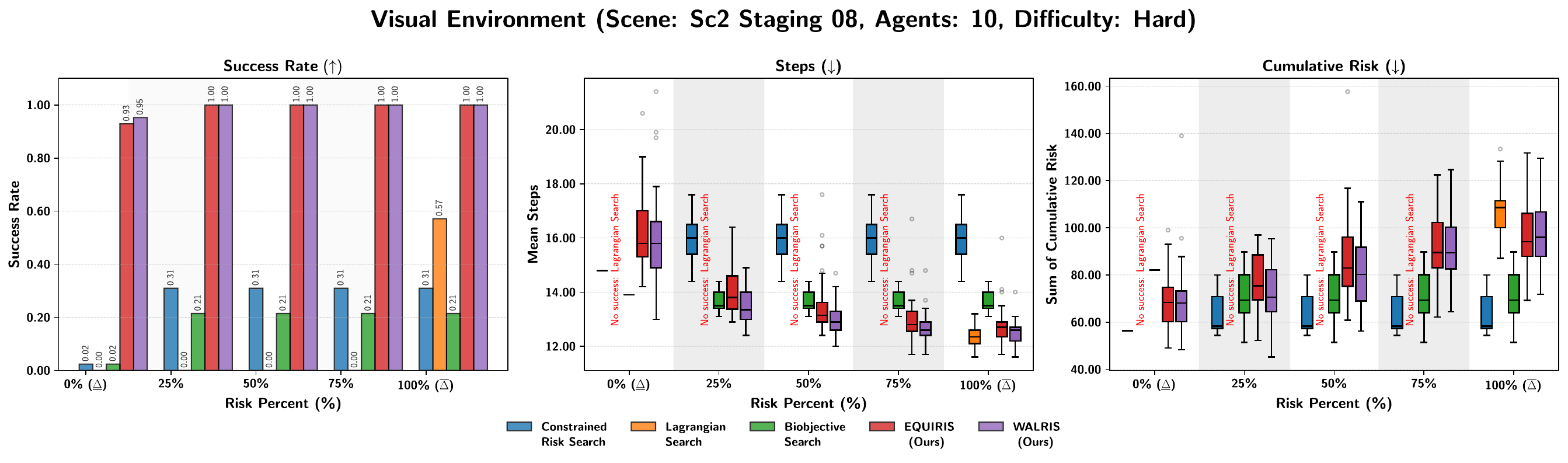}
    \caption{Quantitative results on the visual navigation task (Scene: SC2 Staging 08, Hard difficulty) for 5 agents (top) and 10 agents (bottom). \textbf{Left:} Success rate. \textbf{Center:} Average steps. \textbf{Right:} Total cumulative risk. Our strategies (EQUIRIS and WALRIS) demonstrate superior robustness, maintaining near-perfect success rates even at tight risk budgets ($\Delta \approx 0\%$) where baselines like Lagrangian and Biobjective Search struggle or fail completely. Additionally, the step plots (Center) confirm that our planner effectively exploits relaxed budgets to reduce travel time, scaling reliably to higher agent counts.}
    \label{fig:habitatenv_plot}
\end{figure*}
\section{Experiments}
\label{sec:experiments}

Our experiments evaluate our framework against several baselines along three key questions:

\begin{itemize}[noitemsep, align=left]
    \item[\textbf{Q1 (Adaptability):}] Does the planner effectively leverage varying risk budgets $\Delta$ to trade-off safety and efficiency?
    \item[\textbf{Q2 (Goal Success):}] Can it maintain high success rates for distant goals while enforcing a global risk bound?
    \item[\textbf{Q3 (Scalability):}] Do these advantages persist as the number of agents increases?
\end{itemize}

\textbf{Environments.} We use a simple 2D navigation task and visually rich indoor scenes. For 2D navigation, we use the Central Obstacle map from \citeauthor{feng2025safemultiagentnavigationguided}, with state $s = (x, y) \in \mathbb{R}^2$ and actions $a = (dx, dy) \in [-1, 1]^2$. The per-state risk cost 
\begin{equation}
    C(s) = 
        \begin{cases}
            2 - 2h(s)/r & 0 \le h(s) \le r, \\
            0, & \textnormal{otherwise}
        \end{cases}
    \label{eq:cost_f}
\end{equation}
depends on the distance $h(s)$ to the nearest obstacle boundary, with radius of influence $r$ ($r = 10$ in 2D, $r = 1$ in visual tasks), yielding higher risk near obstacles and zero risk in free space. The GCRL agent is trained with sparse rewards, receiving $-1$ per step. For visual navigation, we adopt four ReplicaCAD scenes \cite{replica19arxiv} in Habitat-Sim \cite{szot2021habitat, habitat19iccv, puig2023habitat3}. Agents receive first-person RGB observations, represented as $32 \times 32$ images from the four cardinal directions concatenated into a panoramic view. The action space matches the 2D setting, and we reuse the same critic architectures from \citeauthor{feng2025safemultiagentnavigationguided}.

\paragraph{Defining risk bounds.}
Since our framework operates under a user-specified global risk budget $\Delta$, we first calibrate a meaningful range per instance to ensure consistent benchmarking. For each instance, we run standard CBS on the learned waypoint graph twice, once to minimize total path length yielding an upper risk bound, $\overline{\Delta}$ (the risk incurred by the shortest-path solution), and once to minimize total risk, yielding a lower risk bound, \underline{$\Delta$}. This defines an instance-specific interval $[\underline{\Delta}, \overline{\Delta}]$. We then evaluate all methods at five risk levels $(0\%, 25\%, 50\%, 75\%, \text{ and } 100\%)$ within this interval, modeling user preferences ranging from strong risk aversion to aggressive efficiency. This calibration is strictly for experimental rigor. In practice, a user could specify a single budget $\Delta$, and our framework can be applied directly without these auxiliary CBS runs. 

\paragraph{Evaluation protocol.}
For each environment (5 total), agent count ($N \in \{5,  10\}$) and difficulty level, we generate 50 problem instances by sampling start-goal pairs at different distances, yielding 1500 distinct problems. We define difficulty using the learned waypoint-graph diameter $D$: easy, medium, and hard instances have start-goal shortest-path distances concentrated around $D/8$, $D/4$, and $D/2$, respectively. Each trial has a time limit of $60 \times N$ seconds. We report three metrics: (i) \textbf{Success Rate} (fraction of collision-free and risk-compliant runs), (ii) \textbf{Average Steps} (conditioned on success), (iii) \textbf{Cumulative Risk} $(\sum_i \rho(\pi_i))$. All experiments use fixed random seeds for problem generation and policy evaluation. Unless stated otherwise, results in the main text use a uniform initial risk allocation at the root. An ablation in the Supplement Sec. A.1 compares this to alternative initializations. For WALRIS, each agent $a_i$ uses a small symmetric neighborhood $\mathcal{N}_i = \{\delta_i - \eta,\, \delta_i,\, \delta_i + \eta\} \cap [\delta_i^{\min}, \delta_i^{\max}]$. We set the neighborhood step size $\eta = 0.05\cdot\Delta$, the price bisection tolerance to $\varepsilon = 10^{-3}$, and cap the number of WALRIS iterations at $K_{\max} = 20$. Finally, to validate practical applicability, we integrated the planner into a ROS2 stack \cite{ros2} to command multiple Crazyflie drones in Gazebo and hardware. Additional details are provided in Supplement Sec. D and videos are available on the project website. \footnote{\url{https://rb-visual-mapf-mers.csail.mit.edu/}}

\paragraph{Baselines.}
We compare against three baselines that all use the same learned waypoint graph. \textbf{Constrained Risk Search}, follows \citeauthor{feng2025safemultiagentnavigationguided} and prunes edges whose predicted risk exceeds a threshold to form a \emph{safe} graph and then runs CBS to minimize path length. \textbf{Lagrangian Search} instead runs CBS on the full graph with a single edge weight given by a linear scalarization of distance and risk using the learned Lagrange multiplier. Finally, \textbf{Bi-Objective Search} (MO-CBS) performs a multi-objective CBS search \cite{mocbs} on the full graph to approximate the (distance, risk) Pareto front, selecting the shortest path solution that satisfies $\Delta$.

\paragraph{Q1: Adaptability to user-specified risk:}
Qualitative analysis (Figures~\ref{fig:single_pointenv_illustration}, \ref{fig:multi_pointenv_illustration}, and~\ref{fig:single_habitatenv_illustration}) confirms that our framework effectively uses the global budget $\Delta$  as a tunable control knob. At the tightest setting ($\Delta = \underline{\Delta}$), the planner generates conservative routes that maintain large clearance from high-risk regions, at the expense of longer paths. As $\Delta$ is relaxed (e.g., $25\%$ of the interval), the planner produces more direct trajectories that pass closer to high-risk regions. This contrasts sharply with Lagrangian and Biobjective Search, which tend to lock into a single behavioral mode regardless of the specific constraint. 

These qualitative patterns are backed by aggregate statistics (Figures~\ref{fig:pointenv_plot} and~\ref{fig:habitatenv_plot}). Our approach (red and purple) exhibits a clear monotonic trend. At low budgets, average steps are high and total cumulative risk is low. As $\Delta$ increases, the trend inverts in a controlled manner. Notably, WALRIS capitalizes on the available budget more effectively than EQUIRIS, consistently finding shorter paths at medium-to-high $\Delta$. This confirms that the market-based mechanism succeeds in ``spending'' the risk resource to purchase efficiency where the greedy scheme falls short.

\paragraph{Q2: Goal success with safety:}
Our framework achieves high success rates while enforcing the global risk bound. Across all risk levels, both strategies outperform the strongest baselines. The main degradation occurs at the tightest budget $(\Delta = \underline{\Delta})$, where the feasible solution space is extremely narrow. Here, EQUIRIS's success rate drops to $74\%$ on the 2D environment and $76\%$ on the visual domain, while WALRIS remains more robust. This behavior is expected as EQUIRIS relies on a single surplus-deficit redistribution, whereas WALRIS can often recover from difficult constraints via iterative price adjustments that explore the allocation space more thoroughly.  Importantly, once $\Delta$ is relaxed slightly above this extreme setting, both EQUIRIS and WALRIS quickly recover to  near-perfect success rates.

\paragraph{Q3: Scalability to more agents:}
We assess scalability by increasing the team size from 5 to 10 agents (Figure~\ref{fig:habitatenv_plot}), consistent with prior work \cite{mocbs}. Despite the increased density of inter-agent conflicts, the characteristic safety-efficiency trade-off persists. As $\Delta$ increases, the average steps monotonically decrease for both group sizes, indicating that the benefits of risk allocation do not collapse under congestion. WALRIS demonstrates superior scalability, maintaining high success rates across $\Delta$ even with 10 agents. While larger teams provide a larger pool of surplus risk for EQUIRIS to harvest, its greedy updates struggle to coordinate tight interactions as effectively as WALRIS's price-mediated negotiation. Overall, these results suggest that the proposed IRA layer scales gracefully, allowing larger teams to exploit global risk budgets as effectively as smaller ones.

\section{Conclusion}
\label{sec:conclusion}
We introduced \problem{} problem to overcome the conservatism of static graph pruning in safe visual navigation. By treating risk as a shared resource ($\Delta$), our approach dynamically allocates local budgets ($\delta_i$) via an Iterative Risk Allocation (IRA) layer. This allows agents to selectively accept risk to shorten paths, while ensuring the global mission remains safe. Extensive experiments in both 2D and photorealistic environments, validated by a hardware-in-the-loop multi-drone demonstration, confirm that our allocation strategies yield superior performance and a tunable, scalable trade-off between mission safety and efficiency.

\section*{Acknowledgments}
We gratefully acknowledge support from the Defence Science and Technology Agency, Singapore. We thank Morgan Schaefer, Shashank Swaminathan, and Kristoff Misquitta for careful proofreading, and Shreya Chaudhary for help with the hardware demonstrations. The views and conclusions contained in this document are those of the authors and should not be interpreted as representing the official policies, either expressed or implied, of the sponsoring organizations or agencies.

\appendix

\section{Additional Experimental Results}
\label{sec:app_extra_results}

\subsection{Ablation: Initial Risk Allocation}
\label{sec:app_ablation}

\begin{figure*}[htp]
    \centering
    \includegraphics[width=\textwidth]{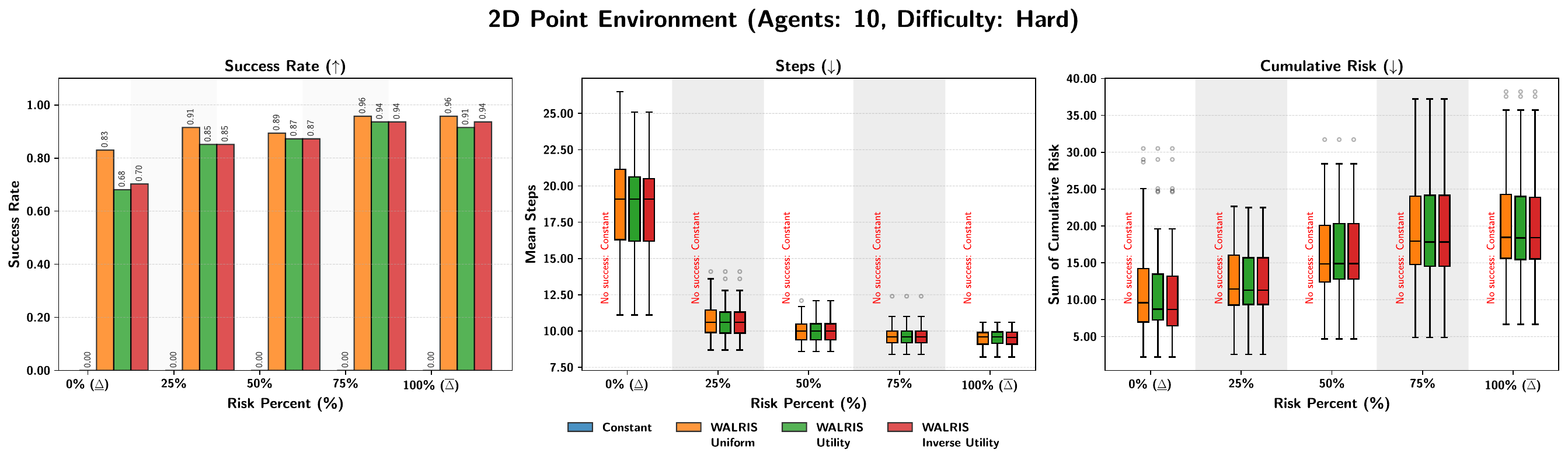}
    \includegraphics[width=\textwidth]{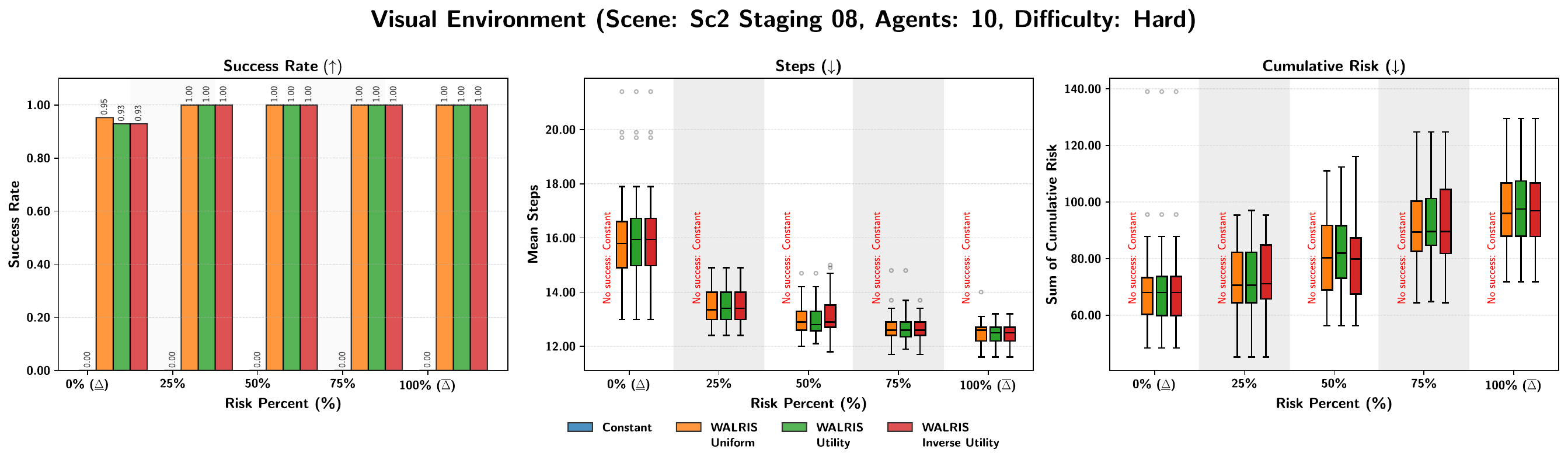}
    \caption{
        \textbf{Ablation Study on Initial Risk Allocation Strategies.}
        Comparison of \textit{Constant}, \textit{Uniform}, \textit{Utility-based}, and \textit{Inverse Utility-based} risk allocation for WALRIS across two distinct environments (10 Agents, Hard Difficulty). The results consistently demonstrate that dynamic allocation strategies (Uniform, Utility and Inverse Utility) have near identical performance and far outperform the rigid Constant approach, especially as risk budgets get tighter. Higher success rates, lower steps and lower risks are better.
    }
    \label{fig:app_full_ablation}
\end{figure*}

Figure~\ref{fig:app_full_ablation} presents an ablation study comparing the static \textbf{Constant} allocation against our three initialization strategies (\textbf{Uniform}, \textbf{Utility-based}, and \textbf{Inverse Utility-based}). The results are consistent across both 2D and visual domains.

The \textbf{Constant} strategy (blue), which fixes budgets at $\Delta/N$ without reallocation, proves extremely brittle. It yields zero success rates across all budgets because it cannot adapt when specific agents encounter hard constraints requiring a larger share of the global budget.

In contrast, all three initialization strategies combined with reallocations demonstrate the robustness of the IRA layer. Regardless of initialization, they maintain high success rates and exhibit the desired adaptive behavior, prioritizing safety (longer paths) at low $\Delta$ and efficiency (shorter paths) as the budget relaxes. The performance gap between the Uniform, Utility, and Inverse-Utility strategies is negligible, suggesting that the system's ability to \emph{reallocate} risk during search is the dominant factor for success, rather than the initial guess. This validates our use of the simple Uniform strategy in the main analysis.

\subsection{Point Environment (Additional Results)}
\label{sec:app_extra_pointenv_results}

\begin{figure*}[htp]
    \centering
    \includegraphics[width=\textwidth]{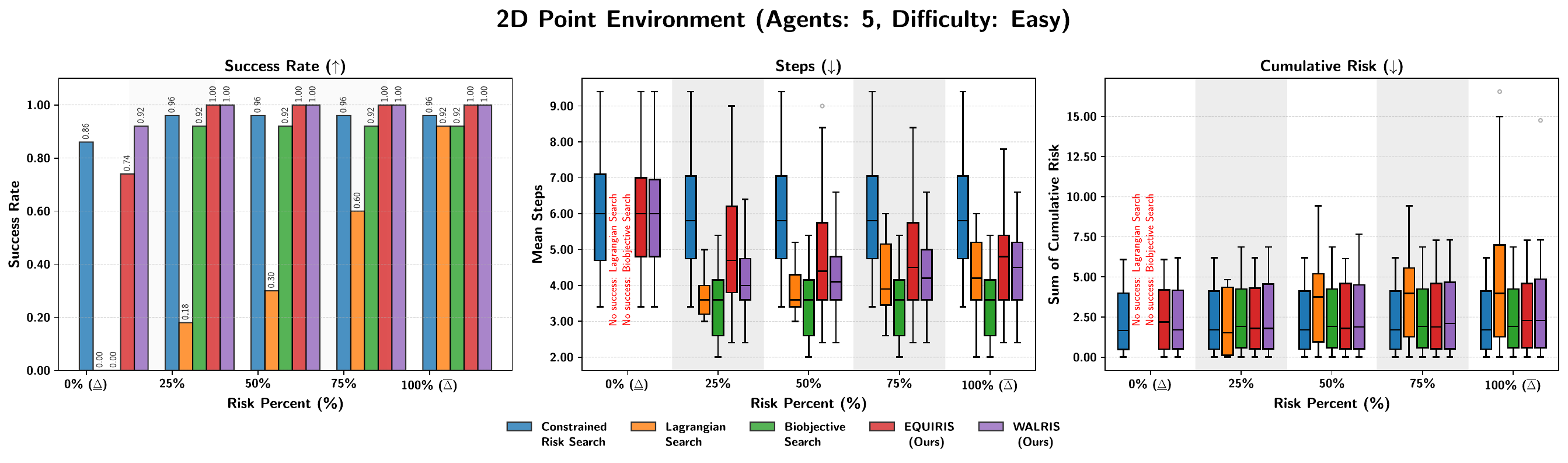}
    \includegraphics[width=\textwidth]{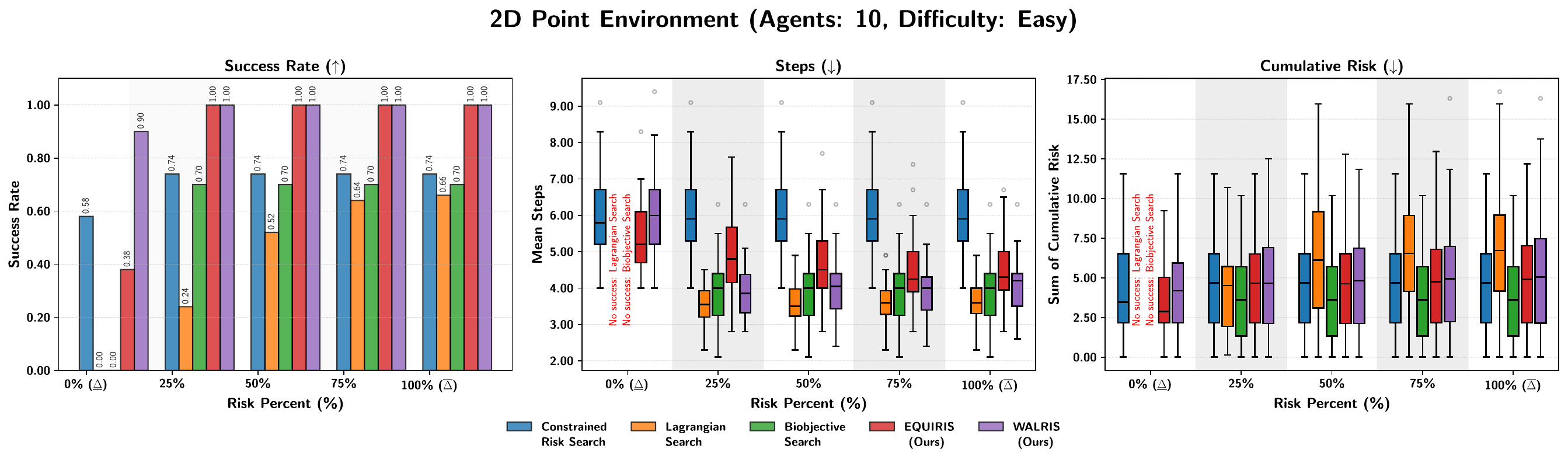}
    \includegraphics[width=\textwidth]{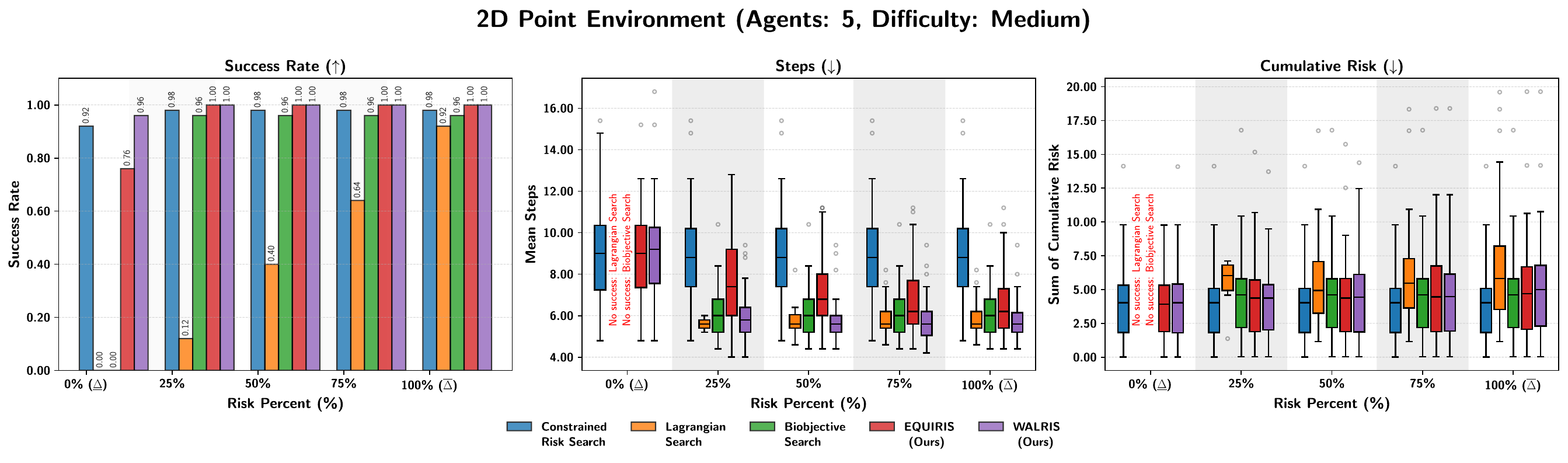}
    \includegraphics[width=\textwidth]{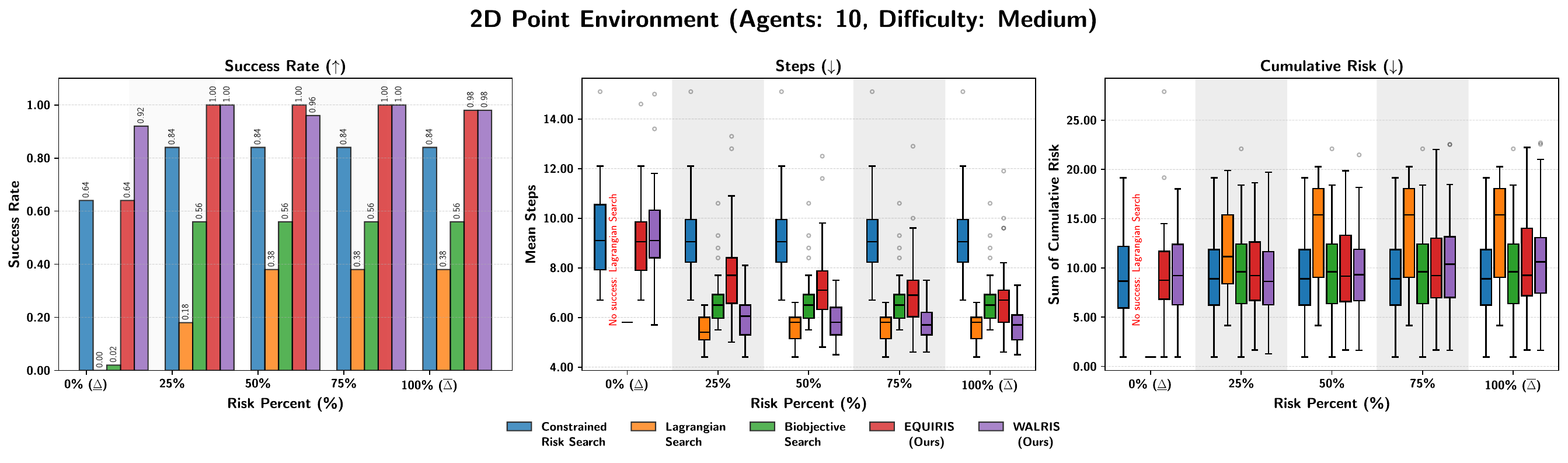}
    \caption{
        Quantitative results for the 2D Point environment comparing all methods across Easy and Medium difficulties for 5 and 10 agents.
    }
    \label{fig:app_pointenv_results}
\end{figure*}

\begin{figure*}[t]
    \centering
    \includegraphics[width=\textwidth]{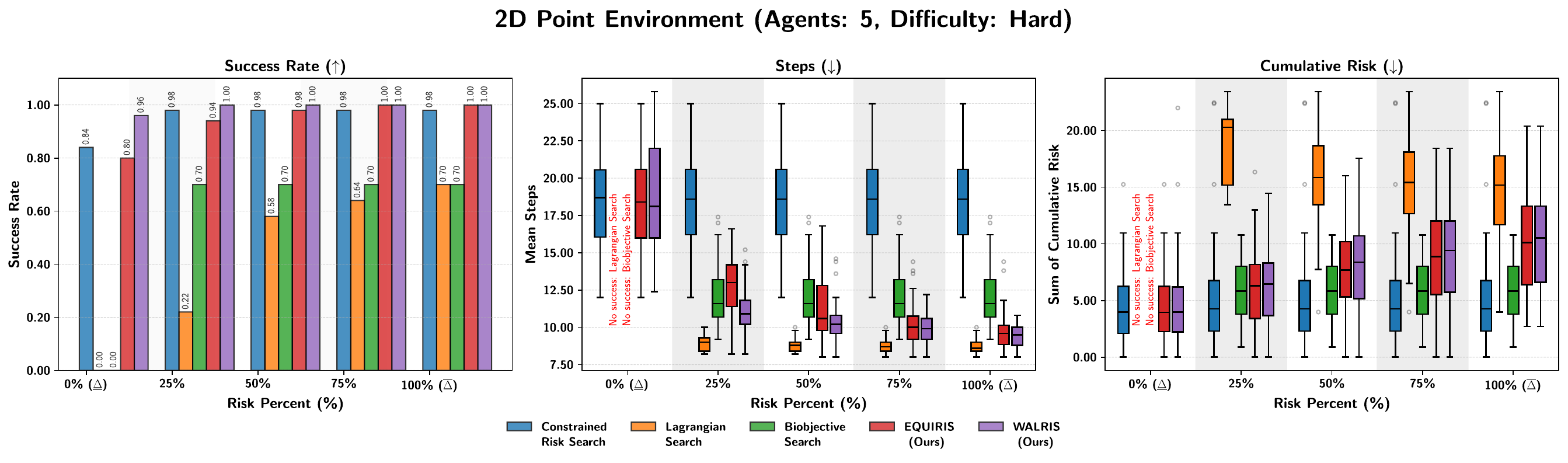}
    \caption{
        Quantitative results for the 2D Point environment comparing all methods across Hard difficulty for 5 agents.
    }
    \label{fig:app_pointenv_results_left}
\end{figure*}

Figures~\ref{fig:app_pointenv_results}-\ref{fig:app_pointenv_results_left} present the additional results for the 2D point environment across \emph{Easy}, \emph{Medium}, and \emph{Hard} difficulties with 5 and 10 agents.

In \emph{Easy} and \emph{Medium} scenarios with 5 agents, the problem is relatively unconstrained as shortest paths naturally avoid high-risk regions. Consequently, all methods achieve high success rates, and the performance curves (steps and risk) remain flat as increasing $\Delta$ offers negligible benefit. However, scaling to 10 agents introduces congestion. Here, we observe a sharp drop in baseline success rates at the tightest risk bounds ($\Delta \approx \underline{\Delta}$) due to dense interactions. Notably, WALRIS maintains superior robustness in these congested, low-budget regimes.

The advantages of our framework become most pronounced in the \emph{Hard} difficulty setting. While our planner maintains high success rates across the full spectrum of $\Delta$, baselines falter. Methods relying on fixed scalarization (Lagrangian Search) or static pruning often exhibit unreliable behavior as they are either unstable at low budgets (high failure rates) or rigid at high budgets (failing to reduce step counts as $\Delta$ increases). In contrast, our approach adapts predictably: incremental relaxations of $\Delta$ reliably translate into shorter, more efficient paths.

\subsection{Habitat Environments (Additional Results)}
\label{sec:app_extra_habitatenv_results}

\begin{figure*}[htp]
    \centering
    \includegraphics[width=\textwidth]{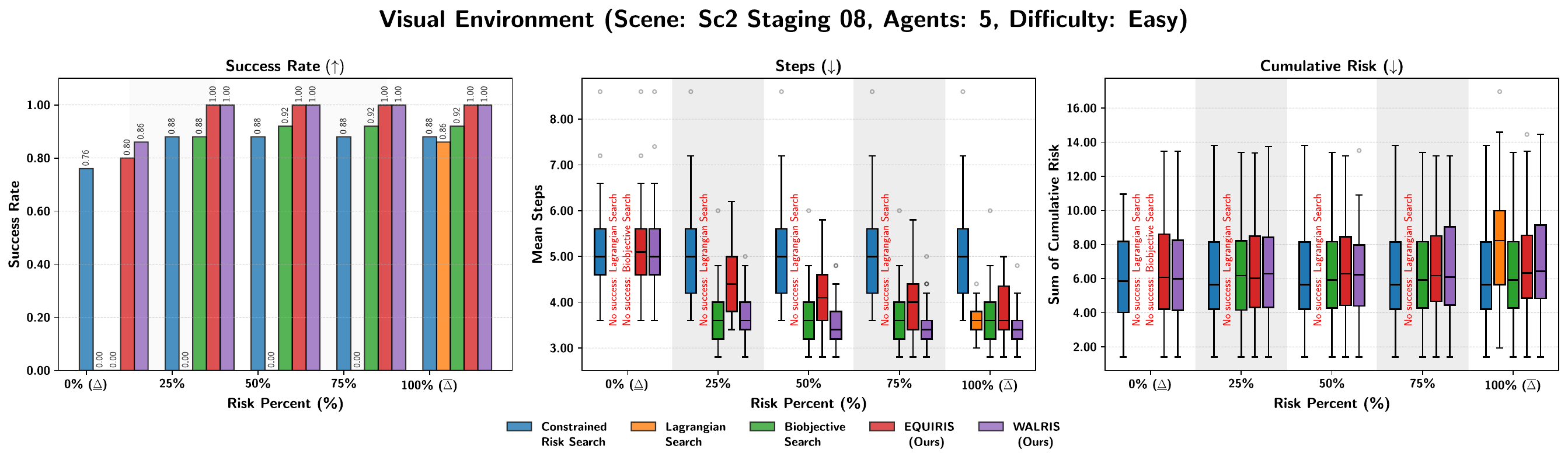}
    \includegraphics[width=\textwidth]{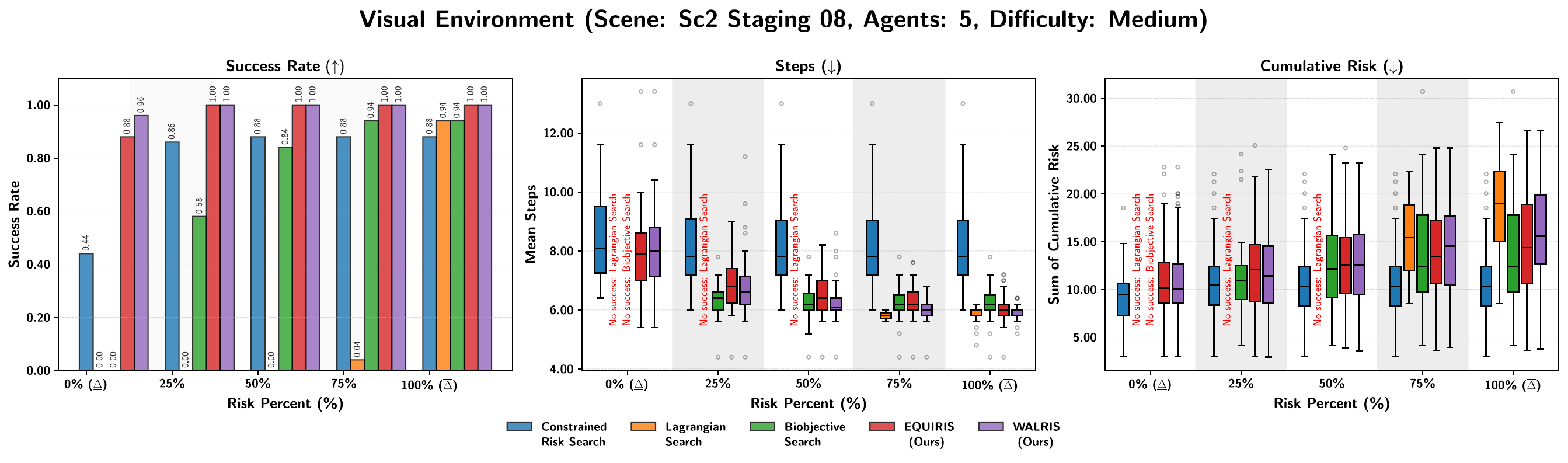}
    \caption{
        Quantitative results for the visual navigation environment on SC2 Staging 08 scene comparing all methods across Easy and Medium difficulties for 5 agents. 
    }
    \label{fig:app_sc2_staging_08_5_results}
\end{figure*}

\begin{figure*}[htp]
    \centering
    \includegraphics[width=\textwidth]{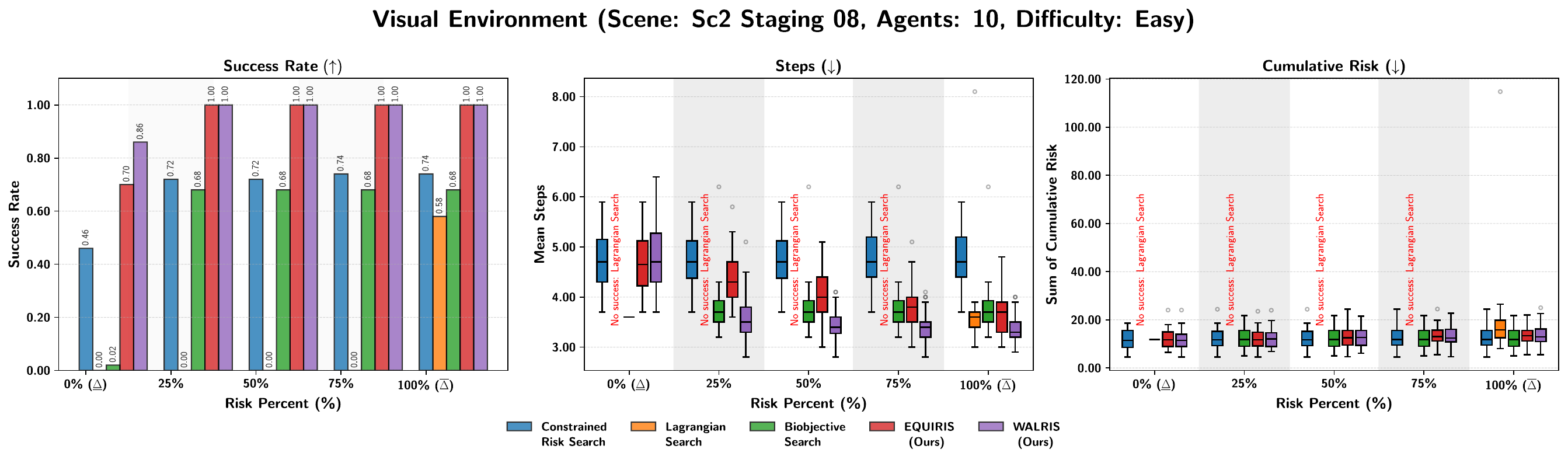}
    \includegraphics[width=\textwidth]{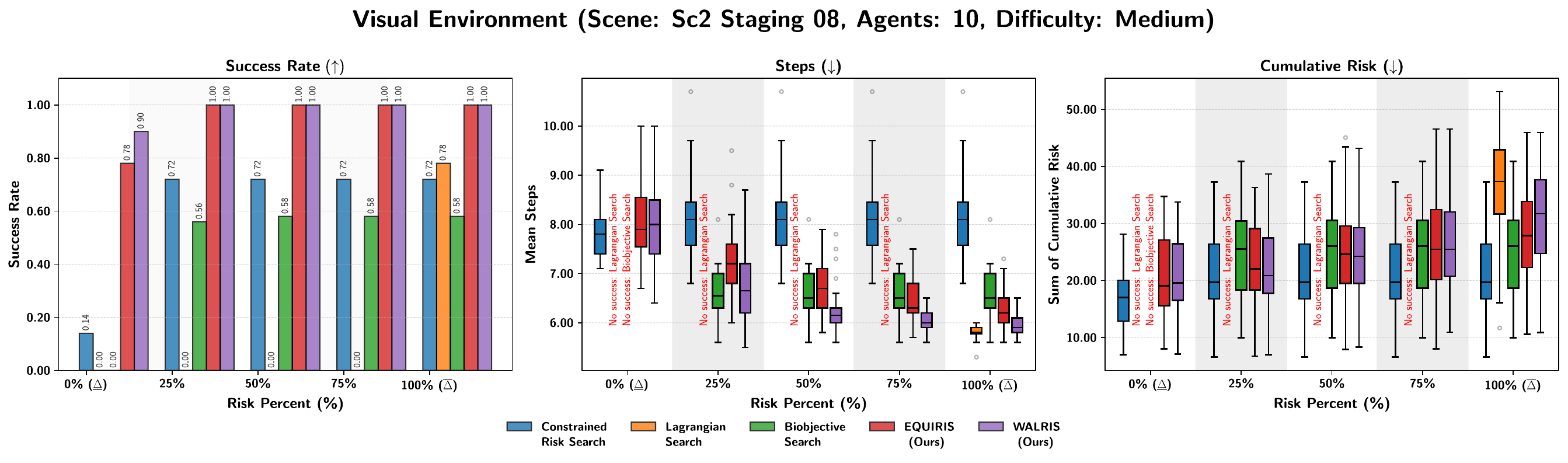}
    \caption{
        Quantitative results for the visual navigation environment on SC2 Staging 08 scene comparing all methods across Easy and Medium difficulties for 10 agents. 
    }
    \label{fig:app_sc2_staging_08_10_results}
\end{figure*}

\begin{figure*}[htp]
    \centering
    \includegraphics[width=\textwidth]{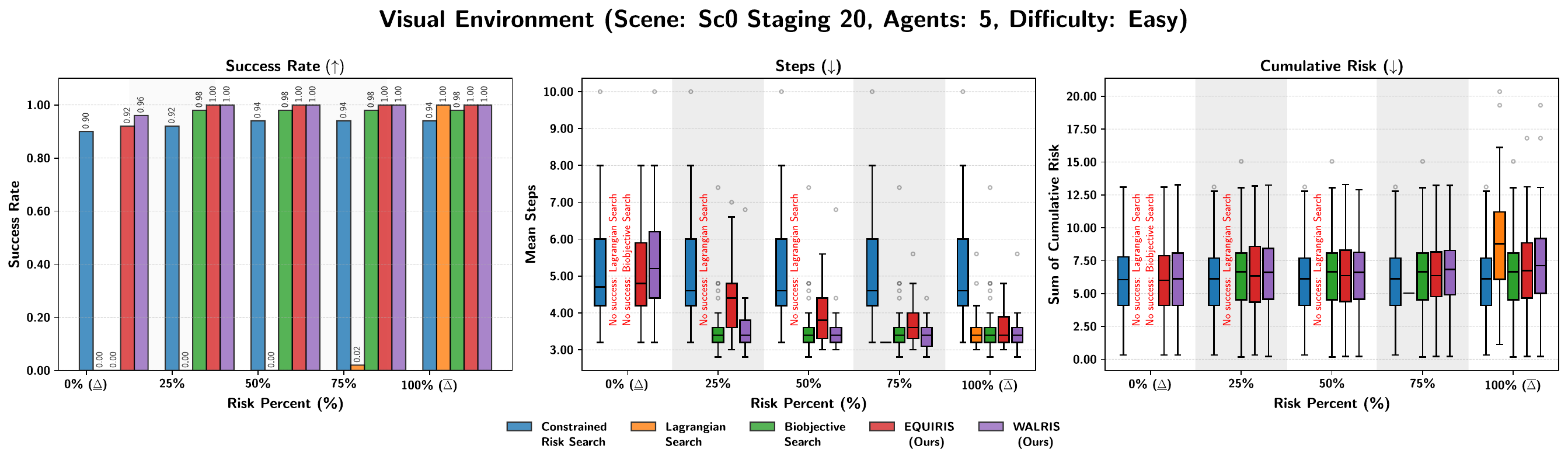}
    \includegraphics[width=\textwidth]{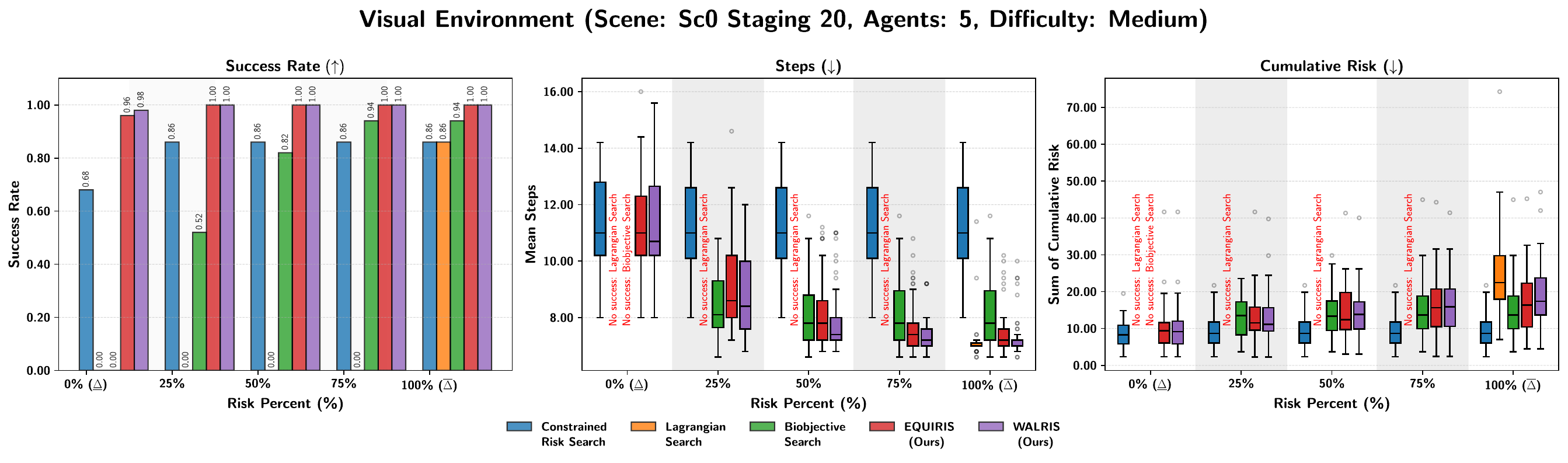}
    \includegraphics[width=\textwidth]{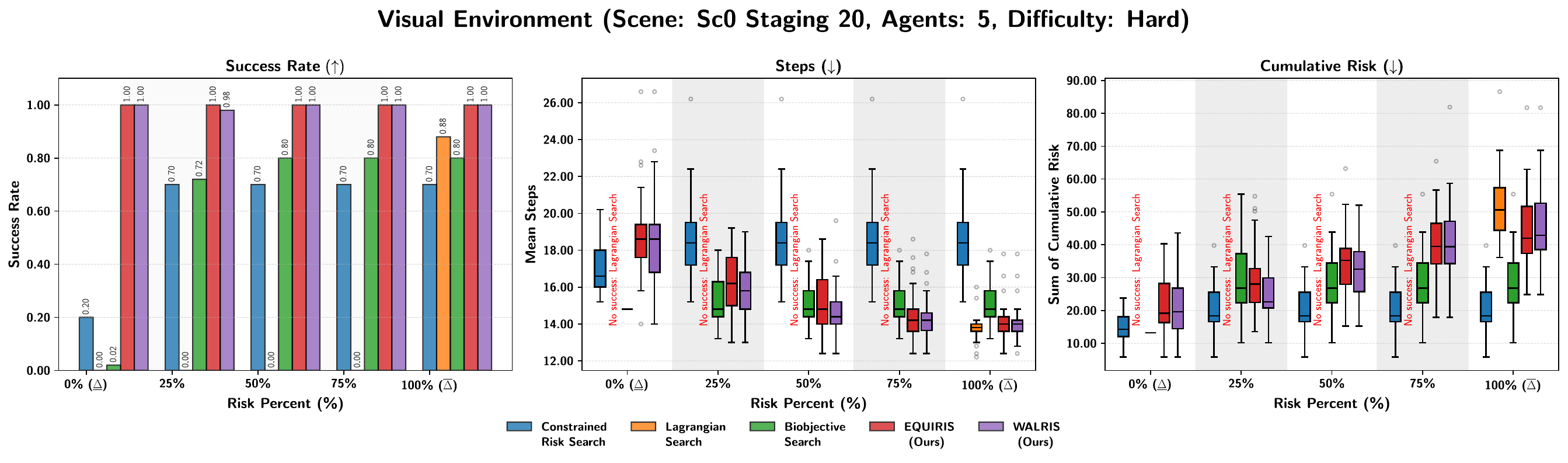}
    \caption{
        Quantitative results for the visual navigation environment on SC0 Staging 20 scene comparing all methods across Easy, Medium and Hard difficulties for 5 agents. 
    }
    \label{fig:app_sc0_staging_20_5_results}
\end{figure*}

\begin{figure*}[htp]
    \centering
    \includegraphics[width=\textwidth]{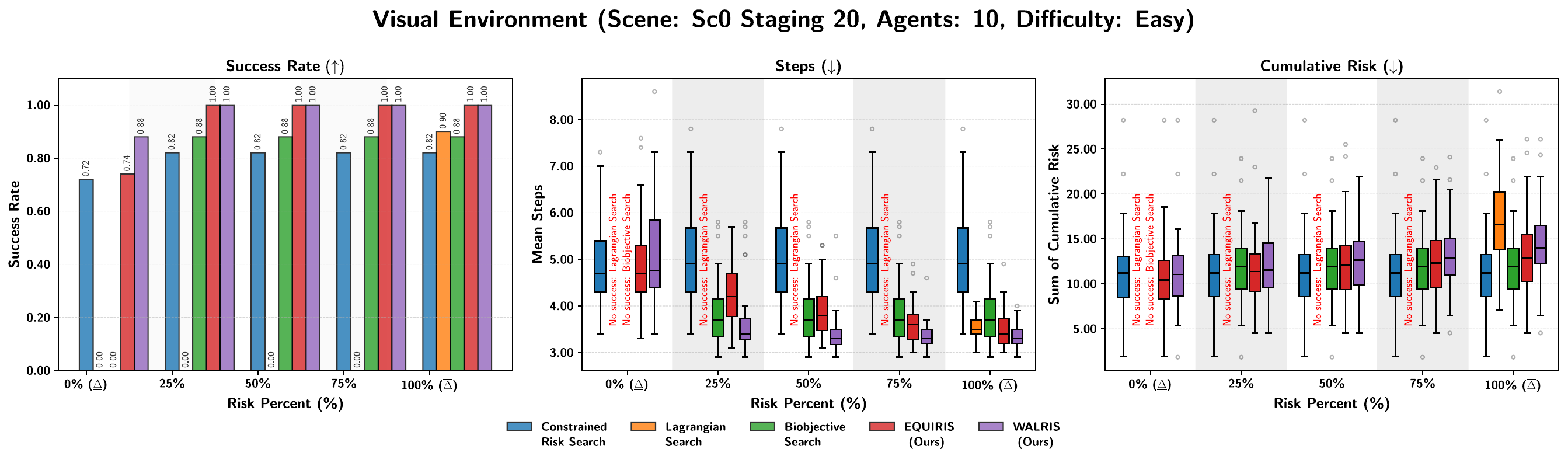}
    \includegraphics[width=\textwidth]{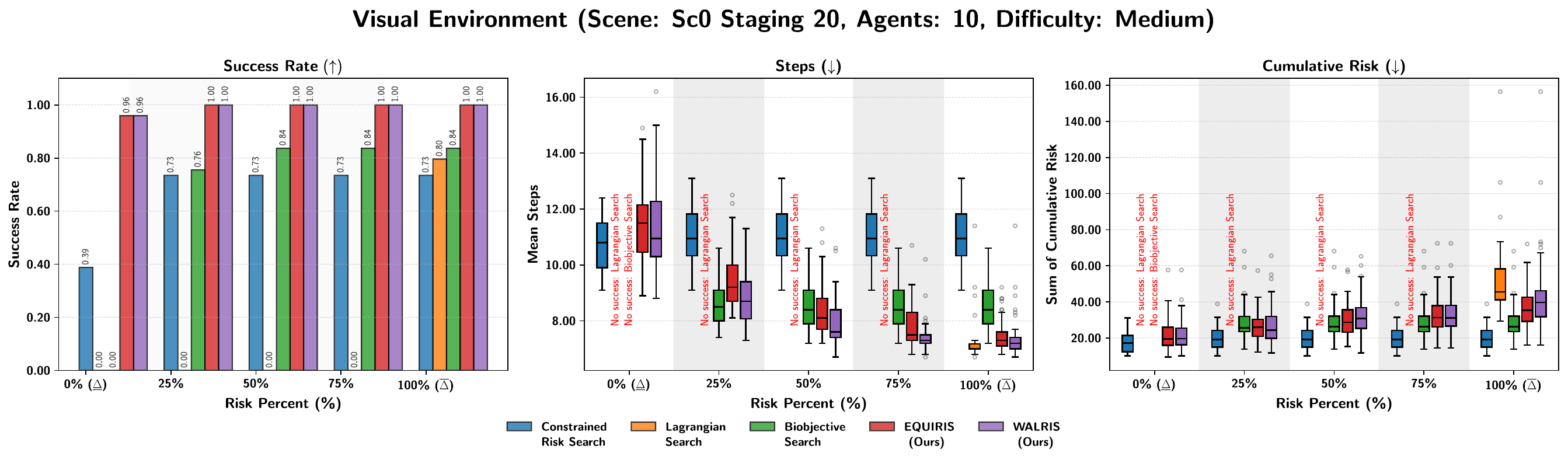}
    \includegraphics[width=\textwidth]{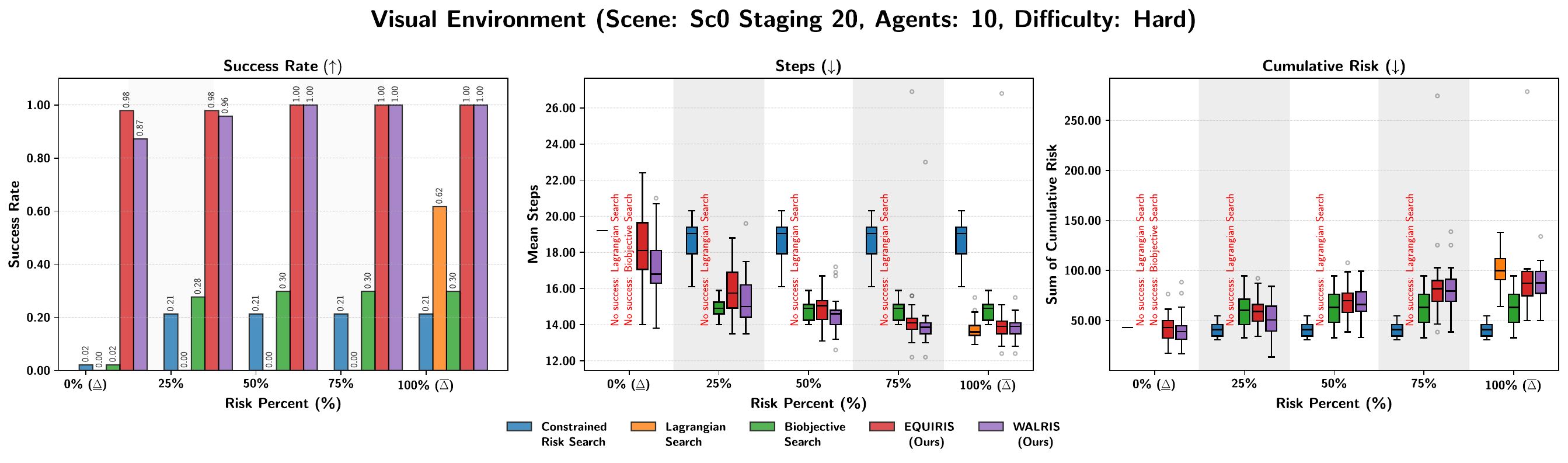}
    \caption{
        Quantitative results for the visual navigation environment on SC0 Staging 20 scene comparing all methods across Easy, Medium and Hard difficulties for 10 agents. 
    }
    \label{fig:app_sc0_staging_20_10_results}
\end{figure*}

\begin{figure*}[htp]
    \centering
    \includegraphics[width=\textwidth]{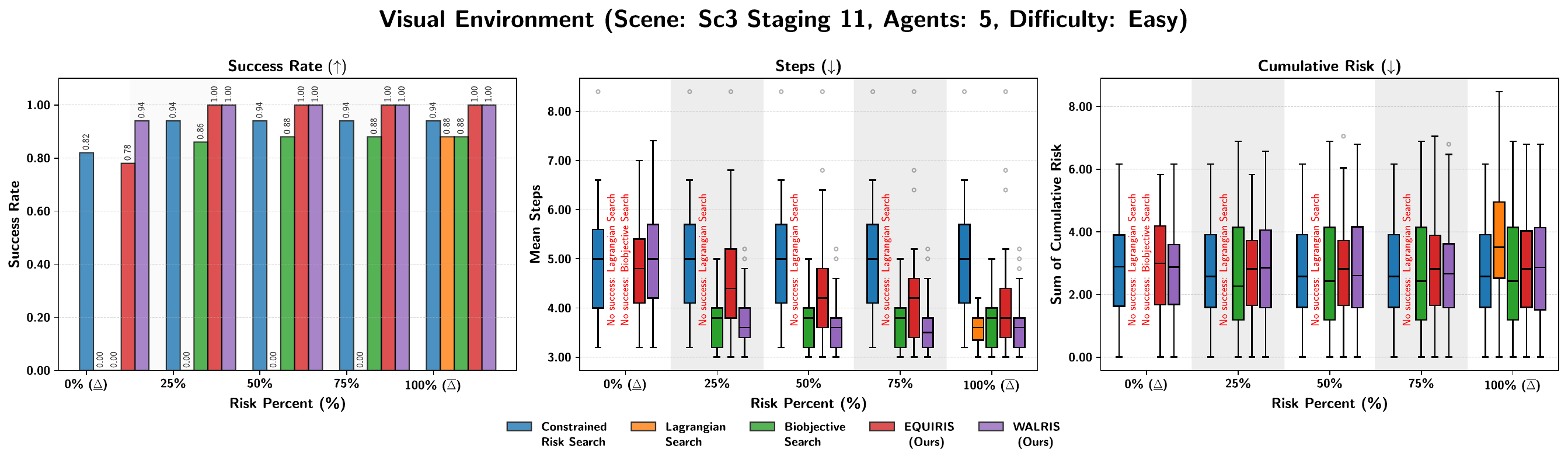}
    \includegraphics[width=\textwidth]{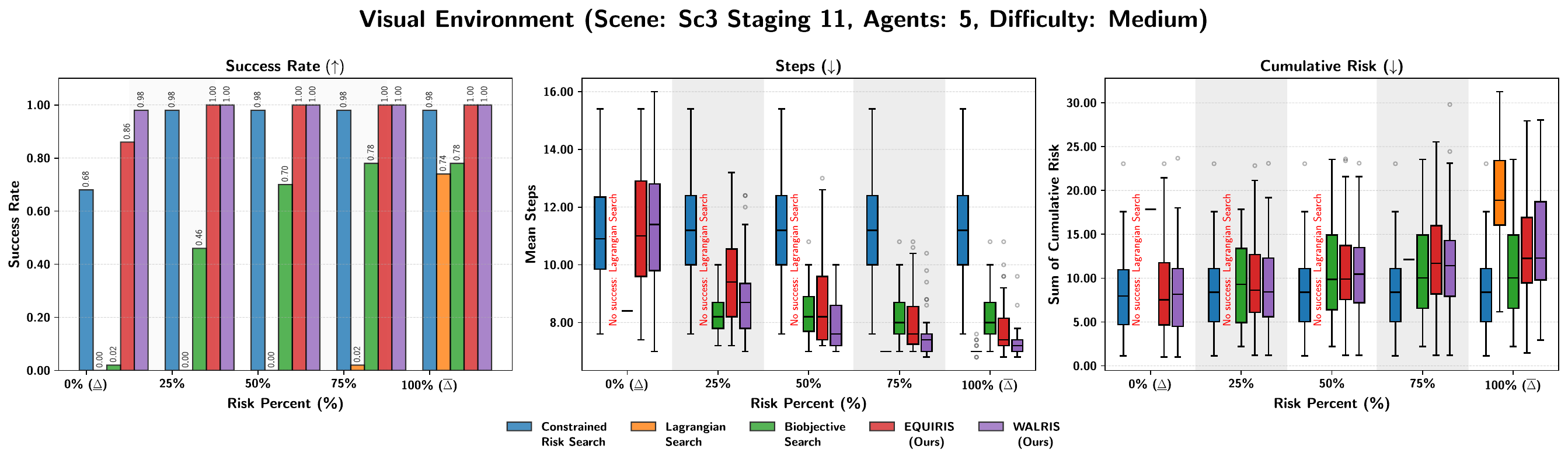}
    \includegraphics[width=\textwidth]{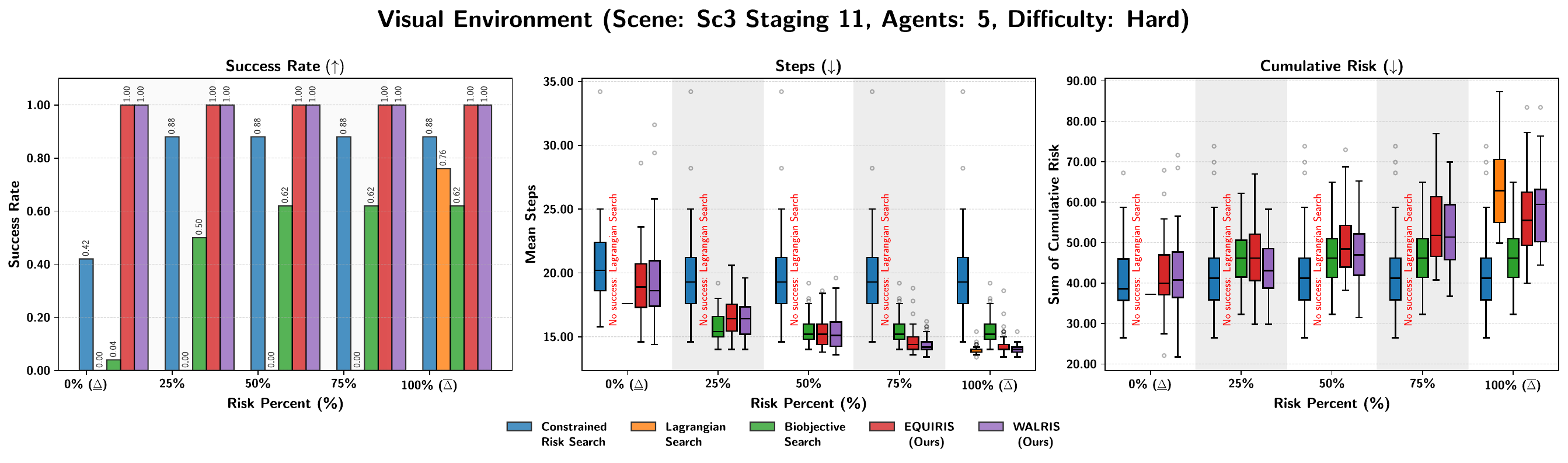}
    \caption{
        Quantitative results for the visual navigation environment on SC3 Staging 11 scene comparing all methods across Easy, Medium and Hard difficulties for 5 agents. 
    }
    \label{fig:app_sc3_staging_11_5_results}
\end{figure*}

\begin{figure*}[htp]
    \centering
    \includegraphics[width=\textwidth]{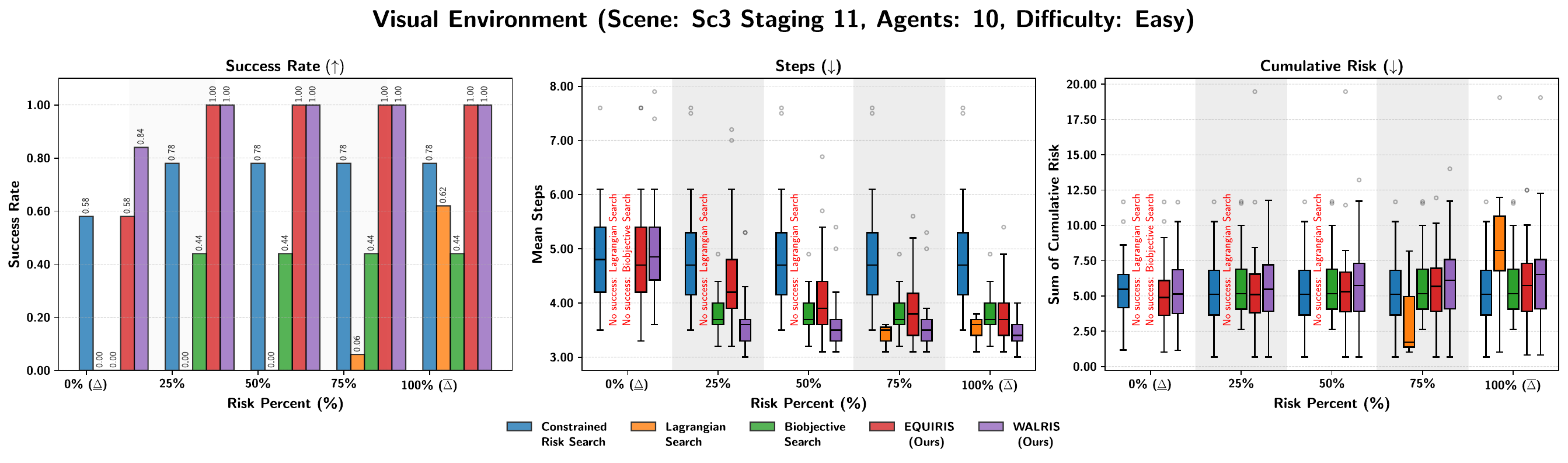}
    \includegraphics[width=\textwidth]{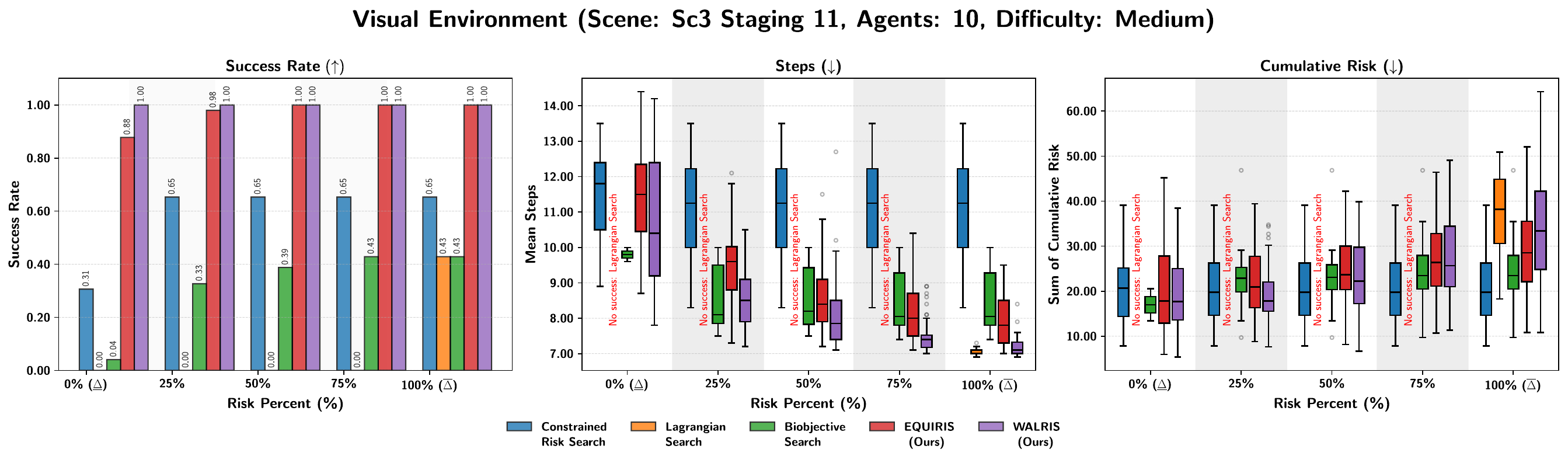}
    \includegraphics[width=\textwidth]{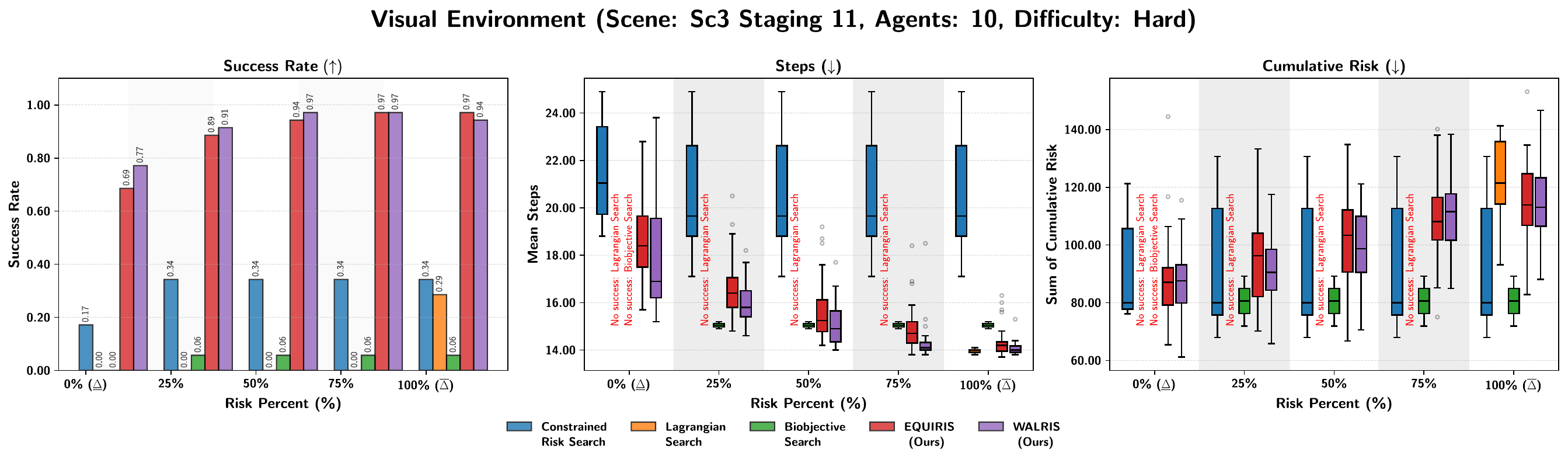}
    \caption{
        Quantitative results for the visual navigation environment on SC3 Staging 11 scene comparing all methods across Easy, Medium and Hard difficulties for 10 agents. 
    }
    \label{fig:app_sc3_staging_11_10_results}
\end{figure*}

\begin{figure*}[htp]
    \centering
    \includegraphics[width=\textwidth]{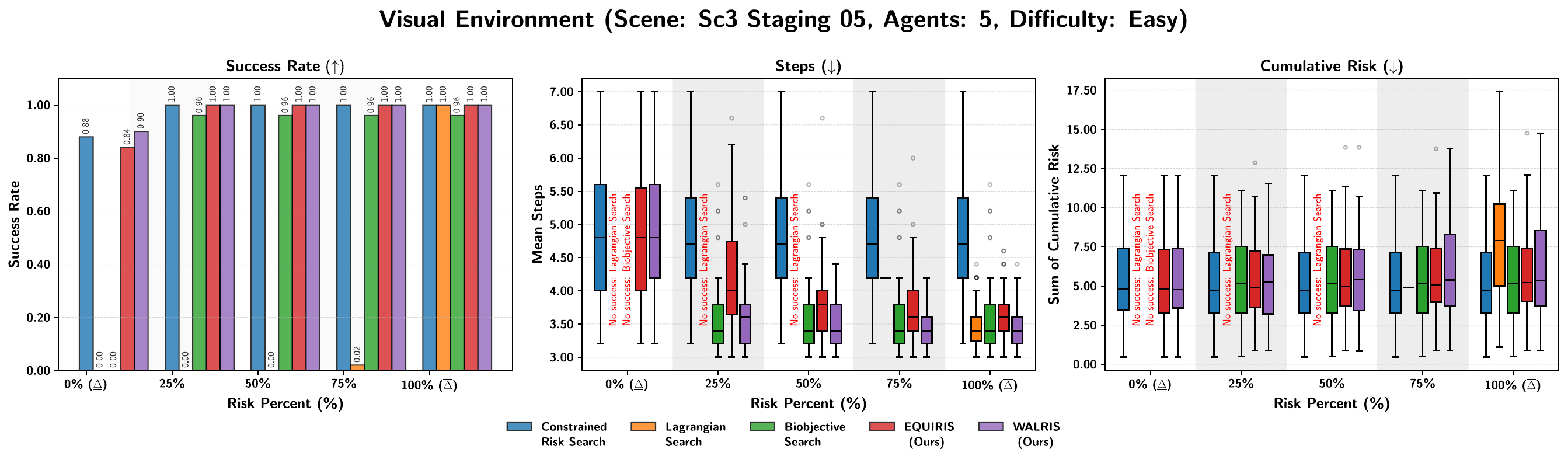}
    \includegraphics[width=\textwidth]{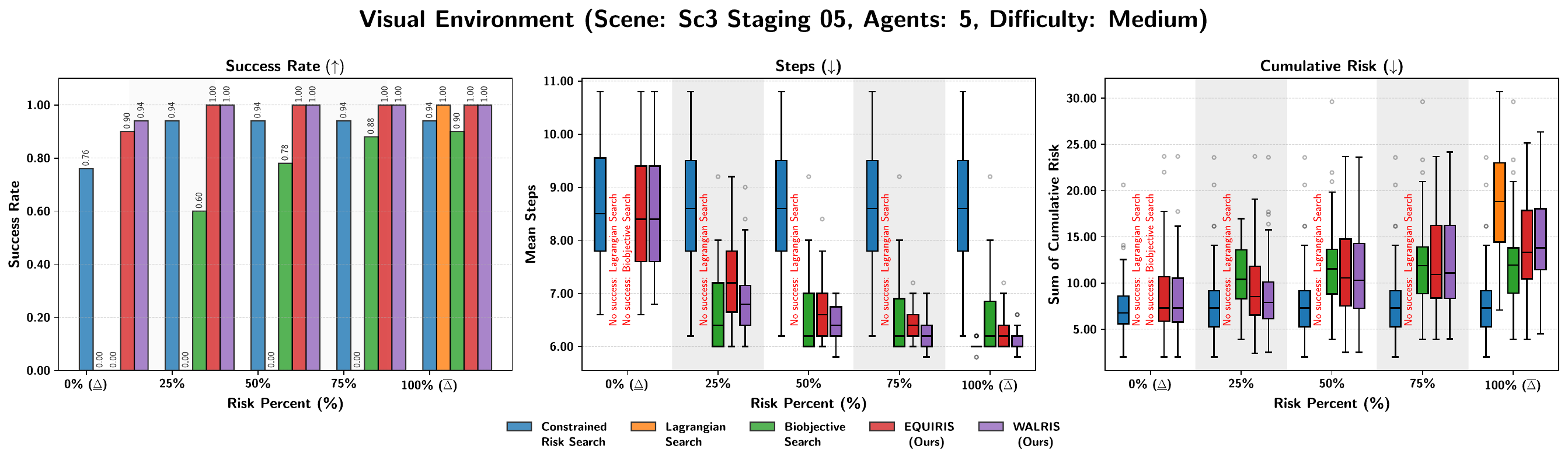}
    \includegraphics[width=\textwidth]{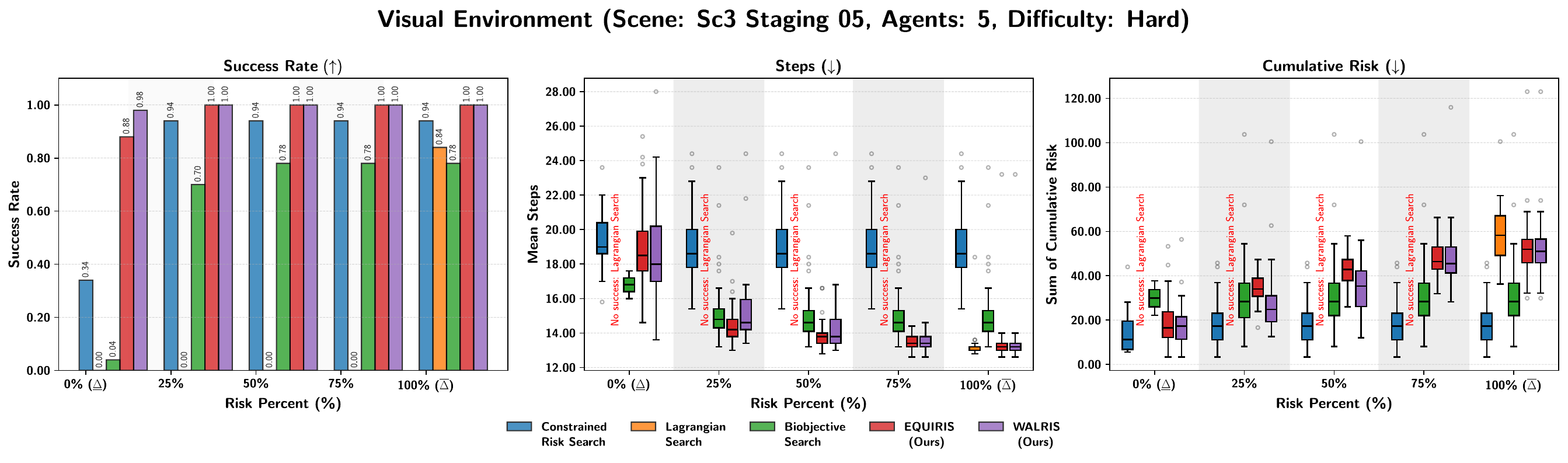}
    \caption{
        Quantitative results for the visual navigation environment on SC3 Staging 05 scene comparing all methods across Easy, Medium and Hard difficulties for 5 agents. 
    }
    \label{fig:app_sc3_staging_05_5_results}
\end{figure*}

\begin{figure*}[htp]
    \centering
    \includegraphics[width=\textwidth]{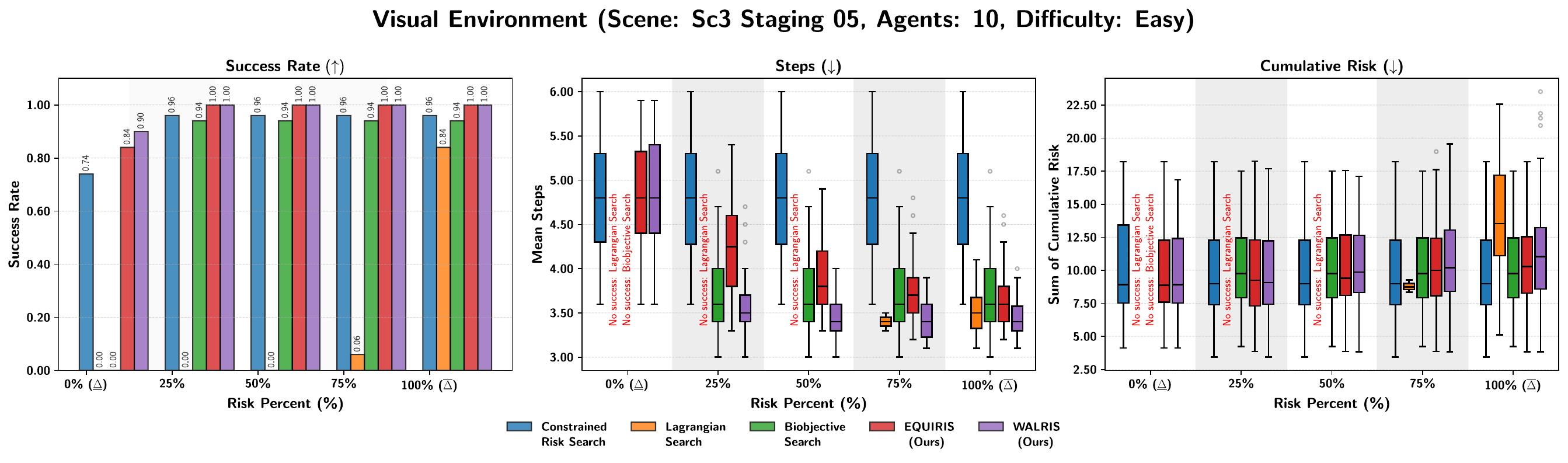}
    \includegraphics[width=\textwidth]{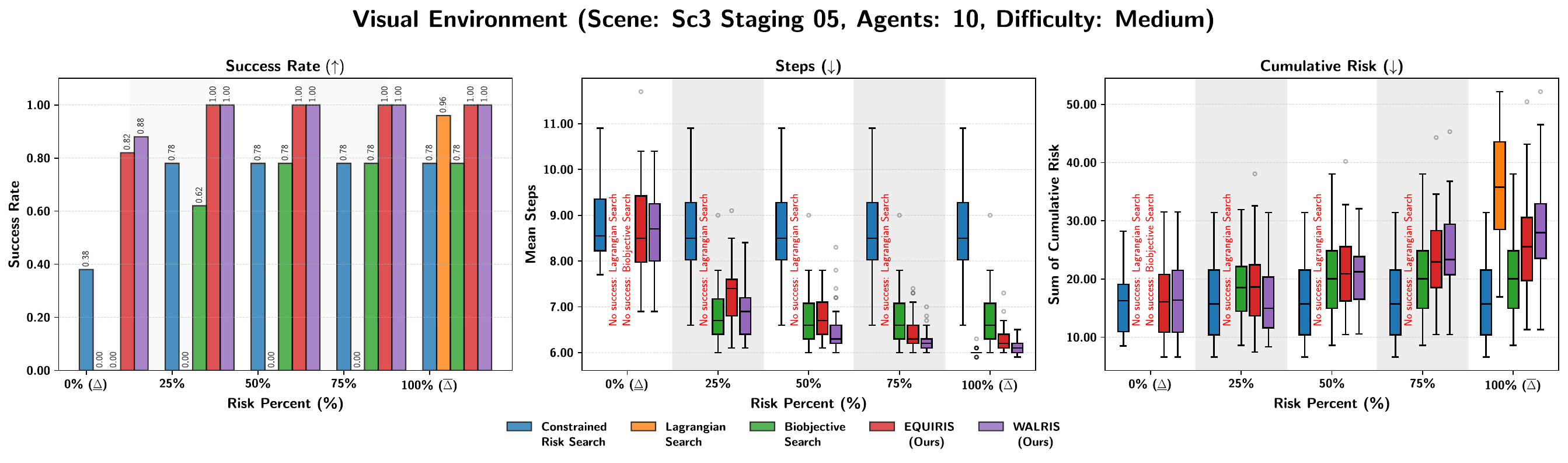}
    \includegraphics[width=\textwidth]{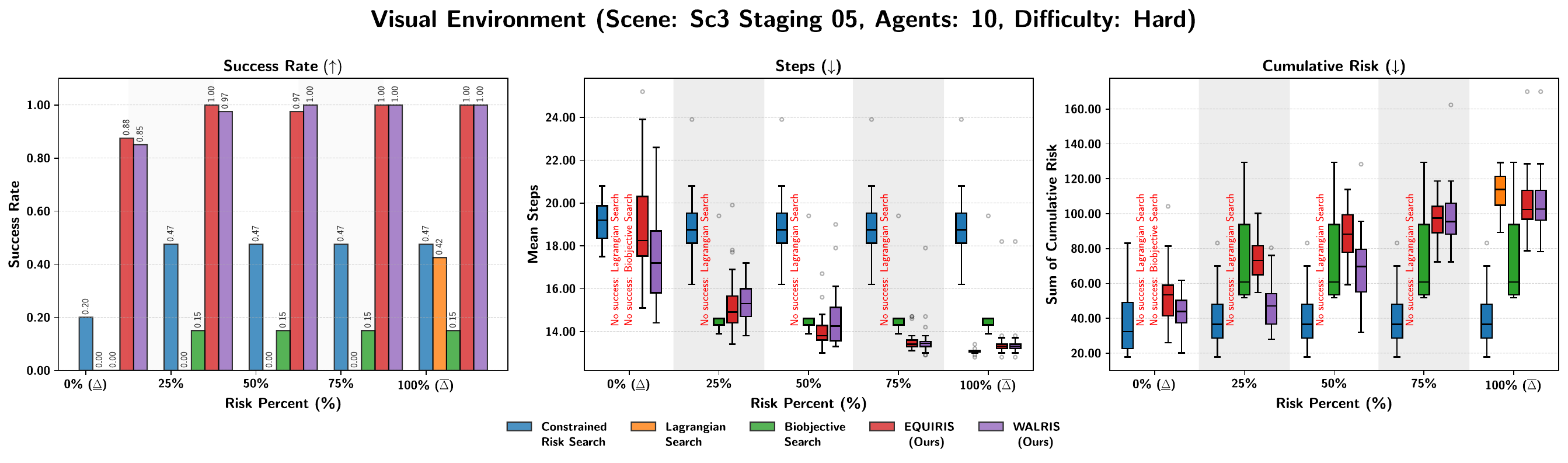}
    \caption{
        Quantitative results for the visual navigation environment on SC3 Staging 05 scene comparing all methods across Easy, Medium and Hard difficulties for 10 agents. 
    }
    \label{fig:app_sc3_staging_05_10_results}
\end{figure*}

Figures~\ref{fig:app_sc2_staging_08_5_results}-\ref{fig:app_sc3_staging_05_10_results} present comprehensive results for the visual navigation tasks across all ReplicaCAD scenes, difficulty levels, and agent counts. The trends align consistently with the main analysis.

Across all scenes, our planner exhibits a characteristic response to the global budget $\Delta$. At tight budgets, it navigates conservatively to minimize cumulative risk; as $\Delta$ relaxes, it systematically trades safety for efficiency, reducing path lengths while respecting the bound. This trade-off is most pronounced in \emph{Hard} settings with 10 agents, where spatial constraints force agents to interact with hazardous regions. Conversely, in \emph{Easy} settings, the cumulative risk curves are flatter, as unconstrained shortest paths naturally avoid hazards.

Our framework (particularly WALRIS) maintains high success rates across nearly all configurations. Performance dips only at the strictest lower bounds ($\Delta \approx \underline{\Delta}$) in the hardest 10-agent scenarios, a regime where the feasible space is extremely small and all planners struggle. Crucially, a slight relaxation of $\Delta$ restores near-perfect success rates without compromising the smooth control over the safety-efficiency trade-off.

Compared to baselines, our planner is the most \emph{responsive} to the user-specified budget. While Constrained Risk and Lagrangian Search often remain static or fail as constraints change, our approach modulates step counts and risk meaningfully with $\Delta$. Biobjective Search offers competitive solutions but suffers from lower robustness at tight bounds. The consistency of these patterns across diverse Habitat scenes confirms that the IRA layer generalizes effectively to different layouts and visual conditions, offering a reliable mechanism for tuning multi-agent coordination.

\section{Additional Algorithmic Discussion}

\subsection{Optimality and Completeness}
\label{sec:app_optimality}

Standard CBS guarantees completeness and optimality for the sum-of-costs objective given a finite graph. However, our introduction of the discrete Iterative Risk Allocation (IRA) layer fundamentally alters these properties. While the underlying \rba{} planner is exact for a \emph{fixed} risk allocation, the mechanism for searching the space of possible allocations is heuristic. Consequently, the full $\Delta$-MAPF planner does not retain strict completeness or optimality guarantees.

\paragraph{EQUIRIS.}
This strategy employs a deterministic, greedy repair. When agents fail, it attempts a specific surplus-deficit transfer based on a fixed ordering of donors. If this single-shot reallocation fails, the node is pruned immediately. Since the algorithm does not backtrack to explore alternative donor orderings or partial transfers, it is incomplete as it may prune a node for which a valid risk distribution technically exists but was not found by the greedy heuristic. Furthermore, because EQUIRIS targets \emph{any} feasible allocation rather than searching for the cost-minimal one, it is inherently suboptimal.

\paragraph{WALRIS.}
This strategy explores a significantly larger portion of the allocation space by optimizing against a global price signal. However, it remains an approximation due to two factors: (1) the local nature of the discrete neighborhood search $\mathcal{N}_i$, and (2) the bounded number of price update iterations. WALRIS does not exhaustively search the full allocation simplex $\{\boldsymbol{\delta} : \sum_i \delta_i \le \Delta\}$. Therefore, it may converge to a local optimum or fail to identify a market-clearing price even when a feasible assignment exists.

In summary, both EQUIRIS and WALRIS are effective heuristic strategies designed to navigate the intractable joint space of combinatorial routing and risk-budget allocation. Future work could explore formulating the allocation step as a Mixed Integer Linear Program (MILP) or using other methods to restore theoretical guarantees at the cost of higher computation time.

\subsection{Additional WALRIS Details}

\paragraph{Price Bracket Initialization.}
WALRIS employs a scalar price $p$ that increases the penalty of risk in each agent's best response. As $p$ increases, agents typically exhibit a preference for safer (lower-risk) paths, so the aggregate risk tends to decrease. To initialize the bracket for bisection, we set $p_\text{low} = 0$ and evaluate the resulting aggregate risk i.e, agents' effectively risk-agnostic paths. If the aggregate risk already satisfies the global budget $\Delta$, then $p = 0$ is the clearing price. Conversely, we set $p_\text{high} = 1$ and iteratively double this value, recalculating the resulting aggregate risk in each step until the total risk falls below $\Delta$. This procedure yields a valid bounding interval on the price, $[p_\text{low}, p_\text{high}]$, for the subsequent steps of the algorithm. If even the minimum-risk options still violate $\Delta$, then no price can satisfy the constraint and the node is deemed infeasible.

\subsection{Limitations and Outlook}
\label{sec:app_limitations}

First, our framework relies on the fidelity of the learned critics. Systematic errors in distance or risk estimation can distort the effective budget and the resulting trade-offs. Consequently, the global bound $\Delta$ constrains the \emph{predicted} cumulative risk on the learned graph, which acts as a proxy for ground-truth safety. Stronger probabilistic guarantees would require integrating uncertainty quantification or calibrating the critics against real-world failure rates.

Second, generalization to unseen environments remains an open challenge. Our current implementation relies on a fixed replay buffer collected within specific scenes. In unseen layouts or under domain shifts, gaps in graph coverage or out-of-distribution critic errors could degrade performance. Deploying this system ``in the wild'' would likely require online graph expansion mechanisms and explicit out-of-distribution detection to handle epistemic uncertainty.

Finally, as a CBS-based approach, our planner inherits exponential worst-case complexity with respect to the number of agents. While we demonstrate robust performance with up to 10 agents in complex visual domains, scaling to the massive fleets typical of abstract MAPF benchmarks remains difficult. A promising direction is to integrate the $\Delta$-MAPF formulation with bounded-suboptimal solvers (e.g., ECBS \cite{barer2014suboptimal}) to manage computational overhead in dense scenarios.

\section{Training Details}

\subsection{Goal-Conditioned RL Training}
\label{sec:app_gcrl_training}

Following \citeauthor{feng2025safemultiagentnavigationguided}, we train a goal-conditioned agent capable of estimating both temporal distance and cumulative risk between states. The training pipeline proceeds in three stages.

\paragraph{Unconstrained Pre-training.}
We initially train an unconstrained goal-conditioned actor-critic. The policy $\mu_\theta(a \mid s, g)$ maximizes sparse rewards for reaching goal $g$. The distance critic $\mathcal{Q}^d_{\theta}$ and risk critic $\mathcal{Q}^c_{\theta}$ are trained off-policy using distributional temporal-difference updates. At this stage, $\mathcal{Q}^d_{\theta}$ learns the shortest-path distance in the state space.

\paragraph{Constrained Fine-tuning.}
To enforce safety, we fine-tune the agent using a Lagrangian actor-critic framework \cite{Ray2019}. We treat the environment risk $C(s)$ as a constraint signal, imposing a soft limit $\bar{c}$. Violations increase a learnable Lagrange multiplier, penalizing the policy for risky behavior. This yields a safety-aware policy $\mu_{\theta}^{\text{safe}}$ and refined critics that accurately reflect the risk-weighted value function, which are subsequently used to construct the planner's graph.

\paragraph{Curriculum Goal Sampling.}
To avoid reliance on ground-truth oracles for goal generation, we employ a self-supervised curriculum. We periodically sample state pairs $(s_i, g_j)$ from the replay buffer and evaluate them using the learned critics. Pairs with predictions matching specific distance or risk curricula are selected for training. These pairs are prioritized for updates using distributional RL \cite{bellemare2017distributional}, ensuring robust critic coverage across a wide range of horizons and risk levels.

\subsection{Hyperparameters}
\label{sec:app_hyperparameters}

Table \ref{tab:hyperparameters} details the hyperparameters used for training. Experiments were conducted on a workstation with a 32-core Intel i9-14900K CPU and an NVIDIA GeForce RTX 4090 GPU. Our implementation is based on the codebase from \citeauthor{feng2025safemultiagentnavigationguided}.

\begin{table}[H]
  \centering
  \begin{tabular}{ll}
    \toprule
    \textbf{Parameter} & \textbf{Value} \\
    \midrule
    Actor Learning Rate           & 1e-5 \\
    Actor Update Interval         & 1 \\
    Critic Learning Rate          & 1e-4 \\
    Risk Critic Learning Rate     & 1e-4 \\
    Distance Critic Bins          & 20 \\
    Risk Critic Bins              & 40 \\
    Targets Update Interval       & 5 \\
    Polyak Update Coefficient     & 0.05 \\
    Initial Lagrange Multiplier   & 0 \\
    Lagrange Learning Rate        & 0.035 \\
    Optimizer                     & Adam \\
    Visual Input Dimensions       & (4, 32, 32, 4) \\
    Replay Buffer Size            & 100,000 \\
    Batch Size                    & 64 \\
    Initial Collect Steps         & 1,000 \\
    Training Iterations           & 600,000 \\
    Neural Network Architecture   & Conv(16, 8, 4) + \\ 
                                  & Conv(32, 4, 4) + \\
                                  & FC(256) \\
    Maximum Episode Steps         & 20 \\
    \bottomrule
  \end{tabular}
  \caption{Hyperparameters and Training Settings for Visual Navigation}
  \label{tab:hyperparameters}
\end{table}

\section{ROS2 / Gazebo and Hardware Demonstration}
\label{sec:app_gazebo_hardware}

To validate the realizability of our risk-bounded plans within a standard robotics control stack, we integrated the planner into a ROS2 pipeline commanding Crazyflie 2.1+ drones via the Crazyswarm2 interface \cite{crazyswarm}. Experiments were conducted in both a high-fidelity Gazebo simulation and a physical motion-capture arena.

\begin{figure*}[t]
    \centering
    \includegraphics[width=0.48\textwidth]{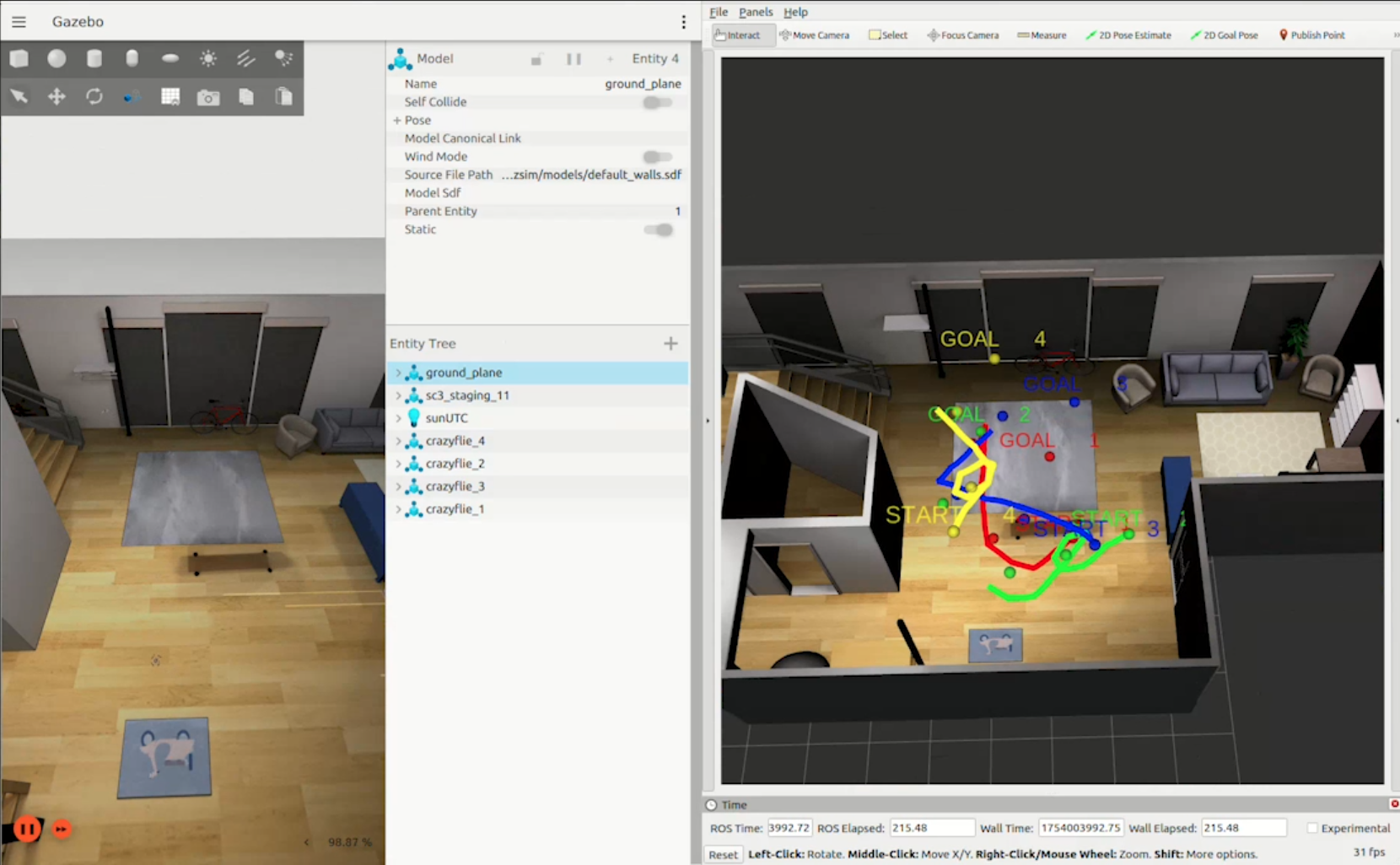}%
    \hfill
    \includegraphics[width=0.48\textwidth]{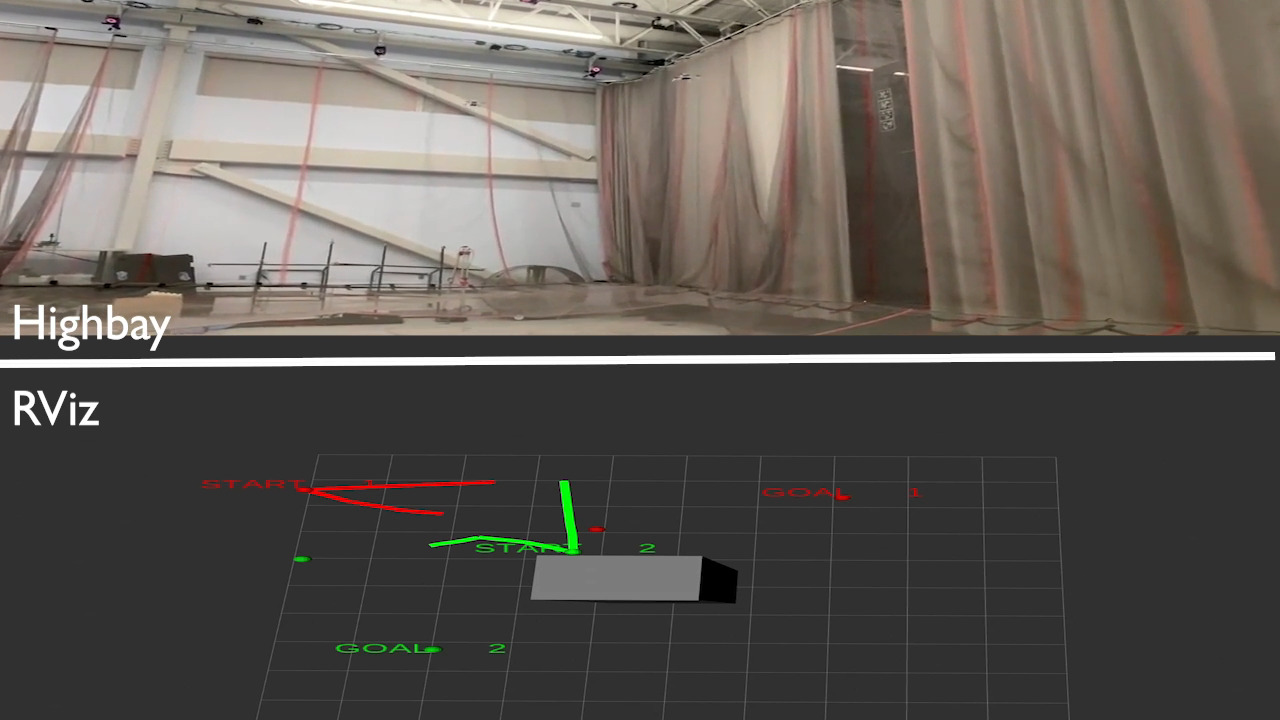}
    \caption{\textbf{ROS2 / Gazebo and hardware demonstrations.}
    \emph{Left:} Multi-agent risk-bounded trajectories executed by simulated nano-quadrotors in Gazebo.
    \emph{Right:} Stills from a hardware experiment with Crazyflie 2.1+ nano-quadrotors in a motion-capture arena using the same planner and ROS2 interface.}
    \label{fig:app_gazebo_hardware}
\end{figure*}

Figure~\ref{fig:app_gazebo_hardware} (left) illustrates a multi-agent simulation in Gazebo where nano-quadrotors navigate around virtual obstacles. Figure~\ref{fig:app_gazebo_hardware} (right) depicts a still from the real-world deployment. In both settings, the drones successfully track the planned trajectories without collisions, confirming effective physical execution.

\begin{figure*}[htp]
    \centering
    \includegraphics[width=\textwidth]{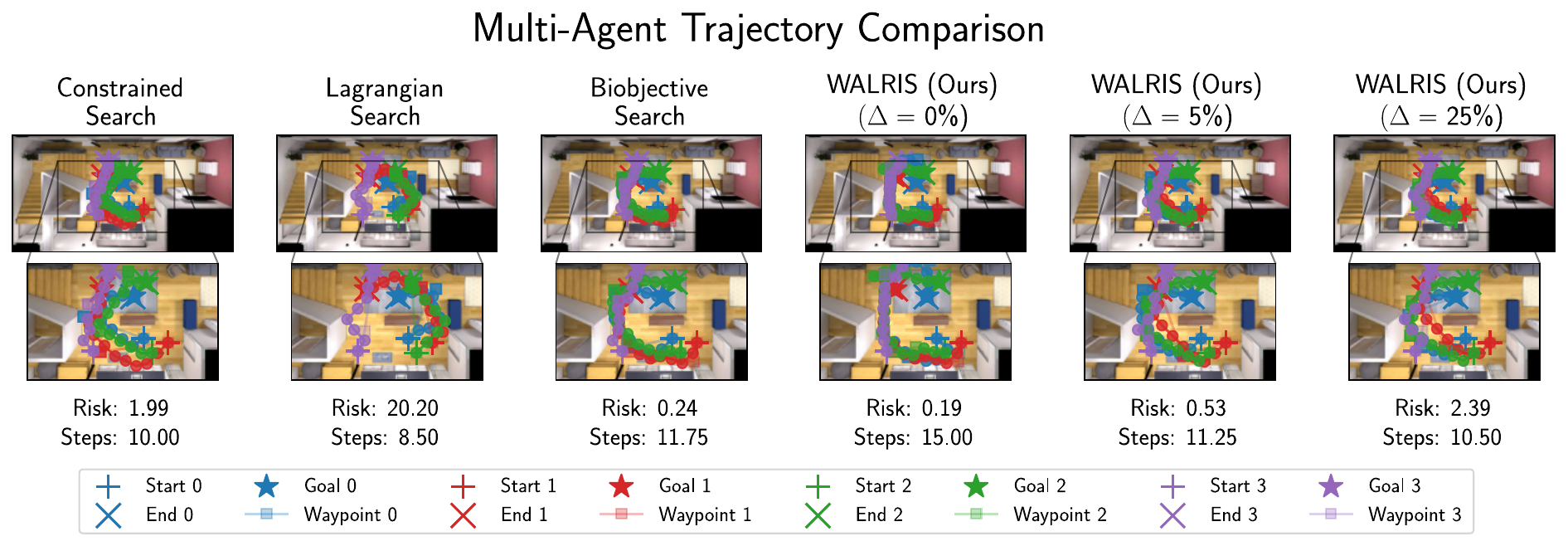}
    \caption{Multi-agent trajectory comparison on the visual navigation task. Cumulative risk  and average step count are annotated below each method. While baselines are locked into static behaviors, either unsafe (Lagrangian, Risk: 20.20) or slightly conservative (Biobjective, Steps: 11.75), our framework (WALRIS) enables a dynamic trade-off. At strict budgets ($\Delta=0\%$), agents coordinate to take wide, safe detours (15.00 steps). As the budget relaxes ($\Delta=25\%$), they utilize the available risk capital to take tighter, more efficient trajectories (10.50 steps), validating the flexibility of the allocation layer.}
    \label{fig:app_multi_habitatenv_illustration}
\end{figure*}

\bibliography{icaps2026}
\end{document}